\documentclass[preprint,12pt]{elsarticle}
\usepackage[utf8]{inputenc}
\usepackage{lineno}
\usepackage{array}
\usepackage[linesnumbered,ruled,vlined]{algorithm2e}
\usepackage{amssymb,amsmath,amsthm,amsfonts}
\usepackage{enumerate}
\usepackage{makecell}
\usepackage{mathtools}
\usepackage{multirow}
\usepackage{slashed}
\usepackage{ulem}
\usepackage{color}
\usepackage{diagbox}
\usepackage{graphicx}
\allowdisplaybreaks[1]
\usepackage{subfigure}
\usepackage{tikz}
\usepackage{indentfirst}
\usepackage{soul}
\usepackage[justification=centering]{caption}
\usepackage{tabularx}
\usepackage{algpseudocode}
\usepackage{adjustbox}
\usepackage{geometry}
\usepackage{booktabs}
\usepackage{longtable}
\usepackage{setspace}
\usepackage{threeparttable}
\usepackage{colortbl}
\usepackage{float}
\usepackage{hyperref}

\newcommand{\circledsmall}[1]{\lower.7ex\hbox{\tikz\draw (0pt, 0pt)%
    circle (.5em) node {\makebox[0.1em][c]{\small#1}};}}

\newcommand{\circledtiny}[1]{\lower.7ex\hbox{\tikz\draw (0pt, 0pt)%
    circle (.3em) node {\makebox[0.1em][c]{\tiny #1}};}}

\journal{Fuzzy Sets and Systems}

\begin{document}

\begin{frontmatter}

\title{A Distance Measure for Random Permutation Set: From the Layer-2 Belief Structure Perspective}

\author[inst1,inst2]{Ruolan Cheng}
\author[inst1,inst3]{Yong Deng\corref{label1}}
\author[inst4]{Serafín Moral}
\author[inst2]{José Ramón Trillo}
\cortext[label1]{Corresponding author.\\ Email address: dengentropy@uestc.edu.cn(Yong Deng).
}

\affiliation[inst1]{organization={Institute of Fundamental and Frontier Science},
organization={University of Electronic Science and Technology of China},
            city={Chengdu},
            postcode={250014},
            country={China}}
\affiliation[inst2]{organization={Andalusian Research Institute in Data Science and Computational Intelligence},
organization={Dept. of Computer Science and AI, University of Granada},
            city={Granada},
            postcode={18071},
            country={Spain}}
\affiliation[inst3]{organization={School of Medicine},
organization={Vanderbilt University},
            city={Nashville},
            postcode={37240},
            country={USA}}
\affiliation[inst4]{organization={Computer Science and Artificial Intelligence, University of Granada},
            city={Granada},
            postcode={18071},
            country={Spain}}

\begin{abstract}
Random permutation set (RPS) is a recently proposed framework designed to represent order-structured uncertain information. Measuring the distance between permutation mass functions is a key research topic in RPS theory (RPST). This paper conducts an in-depth analysis of distances between RPSs from two different perspectives: random finite set (RFS) and transferable belief model (TBM). Adopting the layer-2 belief structure interpretation of RPS, we regard RPST as a refinement of TBM, where the order in the ordered focus set represents qualitative propensity. Starting from the permutation, we introduce a new definition of the cumulative Jaccard index to quantify the similarity between two permutations and further propose a distance measure method for RPSs based on the cumulative Jaccard index matrix. The metric and structural properties of the proposed distance measure are investigated, including the positive definiteness analysis of the cumulative Jaccard index matrix, and a correction scheme is provided. The proposed method has a natural top-weightiness property: inconsistencies between higher-ranked elements tend to result in greater distance values. Two parameters are provided to the decision-maker to adjust the weight and truncation depth. Several numerical examples are used to compare the proposed method with the existing method. The experimental results show that the proposed method not only overcomes the shortcomings of the existing method and is compatible with the Jousselme distance, but also has higher sensitivity and flexibility. 
\end{abstract}


\end{frontmatter}

\section{Introduction}
Information, in reality, almost always carries some form of uncertainty, which may arise from various sources. These include the inherent randomness of natural phenomena or systems, epistemic uncertainty due to insufficient knowledge, fuzzy uncertainty stemming from unclear boundaries of concepts or categories, and uncertainty caused by ambiguity, incompleteness, or complexity. Uncertain information modeling and processing are the theoretical foundations for the next generation of explainable and generalizable artificial intelligence methodologies. So far, numerous mathematical theories have been developed to deal with different types of uncertainty, including probability theory\cite{feller1991introduction}, stochastic processes\cite{doob1942stochastic}, Dempster-Shafer theory\cite{dempster2008upper}, fuzzy set theory\cite{zadeh1965fuzzy}, possibility theory\cite{dubois2001possibility}, Z-number theory\cite{zadeh2011note}, etc. Among these, Dempster-Shafer theory of evidence (DST) has gained significant popularity due to its ability to handle uncertainty without requiring prior probabilities and its flexibility in combining independent and reliable sources of evidence. It expands the basic event space in probability theory to power set space and replaces probability distribution with basic probability assignment (BPA). In exploring the uncertainty of such power set information distribution, researchers introduced Deng entropy\cite{deng2016deng} and provided a novel perspective on the power set\cite{song2021entropic}, suggesting that it encapsulates all potential combinatorial states of basic events. Inspired by this interpretation, Deng\cite{deng2022random} incorporated order into basic events and proposed a novel set concept known as the random permutation set (RPS). Random permutation set theory (RPST) can be regarded as an ordered generalization of DST. By introducing order information, the power set space (combinatorial event space) in DST is further extended to the permutation event space (PES), which encapsulates all potential permutation states of basic events. In RPST, the belief state is given in the form of a permutation mass function (PMF), which is updated by the left orthogonal sum (LOS) and the right orthogonal sum (ROS). 

RPST is a highly potential interpretable information representation framework for general artificial intelligence, which breaks the inherent unorderedness of the classical set and can express higher-dimensional uncertain information. But, we have to admit that it is still a very new theory, having been developed for only three years since its proposal. Existing research mainly focuses on the theoretical exploration of RPS, including entropy\cite{chen2024entropy,zhou2025limit}, distance\cite{chen2023distance}, divergence\cite{chen2023permutation,chen2024symmetric}, negation\cite{tang2024negation}, information dimension\cite{zhao2023information}, combination rules\cite{wang2024new, zhou2024conjunctive,li2025combining}, matrix operations\cite{yang2023matrix}, etc. However, the original paper\cite{deng2022random} on RPST only provided an information representation framework without offering a specific interpretation of this novel form of information distribution. Most of these existing theoretical studies have not addressed this issue. Until recently, Zhou et al. briefly mentioned this in \cite{zhou2023marginalization} and subsequently presented a detailed discussion in \cite{zhou2024conjunctive}, clarifying the meaning of the order introduced in the ordered focal set of RPS from the perspectives of random finite set (RFS) and transferable belief model (TBM) (\cite{zhou2024conjunctive}, Fig. 1). From the perspective of TBM, Zhou et al.\cite{zhou2024conjunctive} proposed the RPS' layer-2 belief structure interpretation, where the order in ordered focal set represents qualitative propensity information, indicating the decision-maker’s belief transfer tendency. Later, based on this interpretation, Deng et al.\cite{deng2024random} introduced a random permutation set reasoning model for classification tasks. Chen et al.\cite{CHEN2025122329} developed a C-means clustering algorithm for RPS by modeling the order as the propensity of samples toward different clusters.


Distance is a binary inconsistency measure that allows the quantifying of notions such as correctness, credibility, or relevance. In DST, the distance between mass functions has been extensively studied\cite{jousselme2012distances} and is usually used as a fundamental criterion for defining the agreement between information sources\cite{schubert2008clustering}, determining discount factors\cite{florea2010dynamic, mercier2008refined}, and evaluating or optimizing algorithms\cite{cuzzolin2004geometry,pichon2010unnormalized}. However, in RPST, the distance measure for RPS is rarely developed. Currently, only Chen et al.\cite{chen2023distance} proposed a distance measure method for RPS by introducing the ordered degree of the permutation events into the weighting matrix $\underline{D}$ in the Jousselme distance\cite{jousselme2001new}. They chose to use multiplication for the ordered degree and the Jaccard index to expand the matrix $\underline{D}$, which changes the positive definiteness of the original matrix, causing their method to no longer satisfy certain metric properties of distance. Moreover, their method does not follow any interpretation of information distribution. Clarifying the specific interpretation of permutation set information distribution is the premise for the theoretical study of RPS and for conducting uncertainty reasoning based on RPS. Under different perspectives, information measures have different behaviors and characteristics. Therefore, how to measure the distance between RPSs under different interpretations remains an urgent problem that needs to be addressed in our practical application of RPS in the future.

This paper focuses on the distance measure between RPSs from the TBM perspective, following the RPS' layer-2 belief structure interpretation. We first deeply analyze and discuss the different behaviors and performances of the distance between RPSs from the two perspectives of RFS and TBM. Under the TBM interpretation, the newly introduced order information in the focal set represents qualitative propensity. It indicates the belief transfer propensity of the decision-maker, characterized by a preference for higher-ranked elements. Based on this characteristic, we propose a new definition of the cumulative Jaccard index to quantify the similarity between two permutations. Then, we employ it as the weighting matrix of the $L_2$ distance and propose a cumulative Jaccard index-based RPS distance measure method. The metric and structural properties of the proposed method are examined, including the positive definiteness of the cumulative Jaccard index matrix, and a corresponding correction scheme is presented. Some numerical examples are used to illustrate the behavioral characteristics and advantages of the proposed method. The proposed method naturally exhibits a top-weightness property, meaning that top-ranked target elements occupy more important positions, which aligns with the view of TBM. The parameter $Orn$ allows the decision-maker to flexibly enhance or weaken this property. Furthermore, the proposed method supports distance measures between RPSs of arbitrary truncation depth. By adjusting the parameter $t$, decision-makers can concentrate solely on the elements of interest according to their preferences or specific application needs. A comparison with the existing method demonstrates that the proposed method not only effectively measures the distance between RPSs following TBM interpretation but also addresses the limitations of the existing method while exhibiting superior flexibility.

The remainder of this paper is organized as follows. Section \ref{Preliminaries} provides a review of the fundamental concepts necessary for distance measures in DST and RPST. Section \ref{Different interpretations} compares and analyzes the RPS distance measure from two different perspectives. Section \ref{proposed} introduces an RPS distance measure method based on the cumulative Jaccard index from the perspective of the TBM. In Section \ref{discussion}, the properties and advantages of the proposed method are illustrated through numerical examples. Finally, the whole paper is summarized in Section \ref{conclusion}.

\section{Preliminaries}\label{Preliminaries}
This section provides the foundational knowledge necessary for the subsequent article. It covers the basic concepts, geometrical interpretation, and distance measures of DST and RPST, among others.

\subsection{Distance and metric spaces}
Since this paper aims to develop an effective distance metric between RPSs, we begin by introducing some axioms of the metric.

$\mathbf{Definition \ 2.1.}$ \textit{(Metric space\cite{o2006metric})} Let $M$ be a set equipped with a metric (or distance function) $d: M\times M\to \mathbb{R}$. An ordered pair $(M, d)$ is called a metric space if and only if $d$ satisfies the following properties for all $(x, y, z)\in M^3$:

$(d1)$ \textit{Nonnegativity: $d(x, y)\ge 0$.}

$(d2)$ \textit{Symmetry: $d(x, y) = d(y, x)$.}

$(d3)$ \textit{Definiteness: $d(x, y) = 0 \Leftrightarrow  x = y$}.

$(d4)$ \textit{Triangle inequality: $d(x, y) \leq d(x, z) + d(y, z), \forall z$.} \\
The property $(d3)$ can be expressed as two separate properties, as follows:

\ \ \ \ \ \ $(d3)'$ \textit{Reflexivity: $d(x, x) = 0$}.

\ \ \ \ \ \ $(d3)''$ \textit{Separability: $d(x, y) = 0 \Rightarrow  x = y$}.

Based on the different subsets of the axioms that $d$ satisfies, the classifications, from weakest to strongest, are \textit{pre-metric, semi-pseudometric, pseudo-metric, quasi-metric, semi-metric, and metric} (\cite{jousselme2012distances}, Table 1). If $d$ satisfies all the properties from $(d1)$ to $(d4)$ above, then $d$ is a (full) metric, where $(d1)$ and $(d3)$ jointly define positive definiteness.

\subsection{Jaccard index}
The Jaccard index\cite{jaccard1912distribution}, the so-called Jaccard similarity coefficient, is a statistical measure introduced by Swiss botanist Paul Jaccard to quantify the similarity between sample sets. It plays a role in the structural distance metrics employed in DST and RPST\cite{jousselme2012distances,jousselme2001new,chen2023distance}.

$\mathbf{Definition \ 2.2.}$ \textit{(Jaccard index\cite{jaccard1912distribution})} Given two sets $A$ and $B$, the Jaccard index measures their similarity by calculating the ratio of the intersection of the two sets to their union. Specifically, defined as
\begin{equation}
  \label{eq1}
J(A,B)=\frac{|A\cap B|}{|A\cup B|}.
\end{equation}
Note that, by definition, the $J (A, B)$ values fall in the range $[0,1]$. A value of $0$ indicates that sets $A$ and $B$ have no common elements, while a value of $1$ indicates that the two sets are identical.

\subsection{Dempster-Shafer theory of evidence}
DST is generally considered an extension of Bayesian probability theory, initially presented by Dempster\cite{dempster2008upper} in the context of statistical inference by introducing multi-valued probability mappings. Later, Shafer\cite{shafer1976mathematical} developed it into a reasoning framework for modeling semantic uncertainty information based on random encoding. It has also inspired reasoning models for epistemic uncertainty, such as the TBM\cite{smets1994transferable}, which no longer relies on a probabilistic perspective.

\subsubsection{Basic representations}
Let $\mathcal{T}$ be a frame of discernment(FoD) consisting of $N$ mutually exclusive objects $\tau_i$, where $i=1, 2,...,N$. Its power set $2^{\mathcal{T}}$ constitutes the event space in DST, which contains $2^N$ possible subsets of $\mathcal{T}$, denoted by
\begin{equation}\label{eq2}
\begin{aligned}
\mathcal{P}(\mathcal{T})&=\{F_0, F_1, F_2, ..., F_l, ..., F_{2^N-1}\}\\
&=\{\emptyset, \{\tau_1\},\{\tau_2\}, ...,\{\tau_N\}, \{\tau_1, \tau_2\}, \{\tau_1, \tau_3\}, ..., \{\tau_1, \tau_N\}, ...,\mathcal{T} \}.
\end{aligned}
\end{equation}

$\mathbf{Definition \ 2.3.}$ \textit{(Mass function\cite{shafer1976mathematical})} The mass function $m$ defined on $\mathcal{T}$ , also referred to as the basic probability assignment (BPA), is a function that assigns a value between 0 and 1 to each subset or proposition within $\mathcal{P}(\mathcal{T})$, subject to the following conditions:
\begin{equation}\label{eq3}
\sum_{F_l\in \mathcal{P}(\mathcal{T})}m(F_l)=1 \quad \text{and} \quad m(\emptyset)=0.
\end{equation}
Notably, in TBM, the condition $m(\emptyset)=0$ is canceled. A non-zero mass for $m(\emptyset)$ is typically associated with conflict or the open-world assumption. In this paper, however, we concentrate exclusively on the closed-world assumption, where $m(\emptyset)=0$ is specified. 

$m(F_l)$ is a measure of belief entirely attributed to the proposition $F_l$ and not to any strict subproposition of $F_l$. When $m(F_l) > 0$, $F_l$ is referred to as a focal set. The collection of all focal sets is denoted by $\mathcal{F}$. Hence, a body of evidence (BoE) can be represented as a triple $<\mathcal{T}, \mathcal{F}, m>$. Each focal set in $\mathcal{F}$ can be labeled with a unique binary code, where the Boolean algebra $\{0,1\}$ is used to indicate the presence or absence of an element of the focal set. The subscript $l$ of $F_l$ corresponds to the decimal representation of this binary code. For example, $F_0 \Leftrightarrow F_{0...000}\Leftrightarrow \emptyset$, $F_5 \Leftrightarrow F_{0...101}\Leftrightarrow \{\tau_1, \tau_3\}$.

\subsubsection{A geometrical interpretation of basic probability assignment}
The geometric interpretation of DST was first introduced by Ronald Mahler in his work \cite{mahler1996combining}, where the BoE is treated as a discrete random variable with a value of $\mathcal{P}(\mathcal{T})$ and a probability distribution of $m$. At this point, Bayesian rules can be applied to explore the relationships within the DST under a random set interpretation. Subsequently, Jousselme et al.\cite{jousselme2001new} disregarded the random aspects of BoEs and defined the distance between them solely based on a "geometric" interpretation, which was later further developed by Cuzzolin \cite{cuzzolin2004geometry, cuzzolin2008geometric}.

Let $\mathbb{R}^{|2^{\mathcal{T}}|}$ be the $2^N$-dimensional Euclidean space with an orthonormal reference frame $\{\mathbf{e}_{F_l}\}_{F_l\in 2^{\mathcal{T}}}$. 
Any vector $\mathbf{v}$ in $\mathbb{R}^{|2^{\mathcal{T}}|}$ can be expressed as 
\begin{equation}\label{eq4}
\mathbf{v}=\sum_{F_l\in\mathcal{P}(\mathcal{T})}v_{F_l}\mathbf{e}_{F_l}=[v_{F_l}, F_l\in\mathcal{P}(\mathcal{T})]',
\end{equation}
where $v_{F_l}\in\mathbb{R}$ is the projection of $\mathbf{v}$ onto the basis vector $\mathbf{e}_{F_l}$.

A BPA $m$ defined on $\mathcal{T}$ is a vector $\mathbf{m}$ of $\mathbb{R}^{|2^{\mathcal{T}}|}$, which satisfies the following conditions:
\begin{equation}\label{eq5}
\sum_{F_l\in\mathcal{P}(\mathcal{T})}v_{F_l}=1\ \ \ and \ \ \ v_{\emptyset}=0,
\end{equation}
where $v_{F_l}\ge 0$ and $v_{F_l}=m(F_l)$. For instance, if there is a BPA $m$ defined on a 3-element FoD, $\mathcal{T}=\{\tau_1, \tau_2, \tau_3\}$ , then $\mathbf{m}$ can be written as:
\begin{equation}
\begin{aligned}
\mathbf{m}&= [v_{F_0}, v_{F_1}, v_{F_2}, v_{F_3}, v_{F_4}, v_{F_5}, v_{F_6}, v_{F_7}]' \\  \notag
& = [v_{\{\emptyset\}}, v_{\{\tau_1\}}, v_{\{\tau_2\}},v_{\{\tau_1,\tau_2\}},v_{\{\tau_3\}},v_{\{\tau_1,\tau_3\}},v_{\{\tau_2,\tau_3\}},v_{\{\tau_1,\tau_2,\tau_3\}}]'.   \notag
\end{aligned}
\end{equation}
In particular, the empty set can be omitted from $\mathbf{m}$ without changing anything. For clarity and convenience, researchers often present them in the sequence $[v_{\{\tau_1\}}, v_{\{\tau_2\}},v_{\{\tau_3\}},v_{\{\tau_1,\tau_2\}},v_{\{\tau_1,\tau_3\}},v_{\{\tau_2,\tau_3\}},v_{\{\tau_1,\tau_2,\tau_3\}}]'$ in practical applications.

\subsubsection{The distance in Dempster-Shafer theory}
Distance measures between two pieces of evidence represented by BPAs have long been a prominent research topic. In \cite{jousselme2012distances}, Jousselme and Maupin conducted a comprehensive survey and analysis of dissimilarity measures within DST. They classified these measures into five categories: the Minkowski family, the inner product family, the Fidelity family, composite distances, and information-based distances. Among these, Jousselme et al.'s distance \cite{jousselme2001new}, a representative of the $L_2$ measure within the Minkowski family, is widely recognized and applied due to its excellent metric properties and structural properties.

$\mathbf{Definition \ 2.4.}$ \textit{(Jousselme et al.'s distance\cite{jousselme2001new})} For two BPAs, $m_1$ and $m_2$, defined on the same FoD $\mathcal{T}$, the distance between them is determined by Jousselme et al. as
\begin{equation}\label{eq6}
d_{BPA}(m_1,m_2)=\sqrt{\frac{1}{2}(\mathbf{m_1}-\mathbf{m_2})\underline{D}(\mathbf{m_1}-\mathbf{m_2})^T}.
\end{equation}
Where $\mathbf{m_1}$ and $\mathbf{m_2}$ are $2^N$-dimensional vectors corresponding to the two BPAs, and $\underline{D}$ is a $2^N\times2^N$ Jacard index matrix. More specifically,

\begin{equation}\label{eq7}
\underline{D}(F_i,F_j) = J(F_i,F_j) = \frac{|F_i\cap F_j|}{|F_i\cup F_j|},\ \ F_i,F_j\in \mathcal{P}(\mathcal{T}).
\end{equation}

Alternatively, Formula (\ref{eq6}) can also be written as

\begin{equation}\label{eq8}
d_{BPA}(m_1,m_2)=\sqrt{\frac{1}{2}(||\mathbf{m_1}||^2+||\mathbf{m_2}||^2-2\langle\mathbf{m_1}, \mathbf{m_2}\rangle)}.
\end{equation}
Here, $\langle\mathbf{m_1}, \mathbf{m_2}\rangle$ refers to the scalar product of vectors $\mathbf{m_1}$ and $\mathbf{m_2}$, which is calculated by
\begin{equation}\label{eq9}
\langle\mathbf{m_1}, \mathbf{m_2}\rangle=\sum_{i=0}^{2^N-1}\sum_{j=0}^{2^N-1}m_1(F_i)m_2(F_j)\frac{|F_i\cap F_j|}{|F_i\cup F_j|},
\end{equation}
with $F_i,F_j\in \mathcal{P}(\mathcal{T})$ for $i, j=0,1,...,2^N-1$. $||\mathbf{m}||^2$ represents the 2-norm of $\mathbf{m}$, given by $||\mathbf{m}||^2=\langle\mathbf{m}, \mathbf{m}\rangle$.

\subsection{Random Permutation Set theory}
RPS is a novel set concept introduced by Deng\cite{deng2022random} in 2022, which shows great potential for representing order-structured uncertain information. Some fundamental definitions are as follows.

\subsubsection{Basic representations}
$\mathbf{Definition \ 2.5.}$ \textit{(Permutation event space\cite{deng2022random})} Given the FoD $\mathcal{T}$ consisting of $N$ mutually exclusive objects $\tau_i$, where $i = 1, 2, \dots, N$. The permutation event space in RPST is the set of all possible permutations of the elements in $\mathcal{T}$, denoted as $\mathcal{PES}(\mathcal{T})$.
\begin{equation}\label{eq10}
\begin{aligned}
\mathcal{PES}(\mathcal{T})&=\{F_0^1, F_1^1, F_2^1, ..., F_l^o, ..., F_{2^N-1}^{N!}\}\\
&=\{\emptyset, (\tau_1),(\tau_2), ...,(\tau_N), (\tau_1, \tau_2), (\tau_2, \tau_1), ..., (\tau_{N-1}, \tau_N),  \\
&\ \ \ \ \  (\tau_{N}, \tau_{N-1}), ..., (\tau_1, \tau_2, ..., \tau_N), ..., (\tau_N, \tau_{N-1}, ..., \tau_1) \}
\end{aligned}
\end{equation}
Where the cardinality of the set $\mathcal{PES}(\mathcal{T})$ is given by $\Delta=|\mathcal{PES}(\mathcal{T})|=\sum_{k=0}^{N}P(N, k)=\sum_{k=0}^{N}\frac{N!}{(N-k)!}$. Note that $(\cdot)$ is used to represent permutation events to distinguish them from combinatorial events $\{\cdot\}$ in the $\mathcal{P}(\mathcal{T})$.

$\mathbf{Definition \ 2.6.}$ \textit{(Permutation mass function\cite{deng2022random})} A permutation mass function is a mapping from any ordered set in $\mathcal{PES}(\mathcal{T})$ to $[0, 1]$, denoted as $Perm: \mathcal{PES}(\mathcal{T}) \rightarrow [0, 1]$, and is subject to the conditions:
\begin{equation}\label{eq11}
\sum_{F_l^o\in \mathcal{PES}(\mathcal{T})}Perm(F_l^o)=1\quad \text{and} \quad  Perm(F_0^1)=0.
\end{equation}
$F_l^o$ is called an ordered focal set when $m(F_l^o) > 0$. The set of all ordered focal sets is denoted by $\mathcal{OF}$. The body of evidence in RPST can be expressed as a triple $<\mathcal{T}, \mathcal{OF}, Perm>$. Any ordered focal set $F_l^o$ in $\mathcal{OF}$ can be uniquely identified by a 2-tuple $(l,o)$, where $l$ is interpreted as it does in DST, and $o$ represents the order of $F_l$, ranging from $1$ to $|F_l|!$. Zhou et al. provided a reversible conversion algorithm between them in \cite{zhou2024conjunctive}, enabling more efficient assignment of $Perm$ in engineering programming.
Table \ref{tab1} shows the encoding of ordered focal sets in RPST for an $N$-element FoD.

\begin{table}[!htbp]\small
    \begin{center}
         \caption{Encoding of ordered focal sets in RPST}
            \begin{tabular}{ccccccc}
             \toprule
              $\emptyset$ &  $(\tau_1)$ &  $(\tau_2)$ &  $(\tau_1\tau_2)$ &  $(\tau_2\tau_1)$  &  $(\tau_3)$  &  $(\tau_1\tau_3)$  \\
             \hline
             $F_{0}^1$ & $F_{1}^1$ & $F_{2}^1$ & $F_{3}^1$ & $F_{3}^2$ & $F_{4}^1$ & $F_{5}^1$ \\
             \hline
             $(\tau_3\tau_1)$& $(\tau_2\tau_3)$ &  $(\tau_3\tau_2)$& $(\tau_1\tau_2\tau_3)$ &  $(\tau_1\tau_3\tau_2)$ &$(\tau_2\tau_1\tau_3)$  & $(\tau_2\tau_3\tau_1)$ \\
             \hline
             $F_{5}^2$  &$F_{6}^1$ & $F_{6}^2$ & $F_{7}^1$ & $F_{7}^2$ & $F_{7}^3$ & $F_{7}^4$ \\
             \hline
             $(\tau_3\tau_1\tau_2)$  &$(\tau_3\tau_2\tau_1)$ &...&$(\tau_1\tau_2...\tau_N)$&$(\tau_1...\tau_N\tau_{N-1})$&...&$(\tau_N\tau_{N-1}...\tau_1)$\\
             \hline
             $F_{7}^5$ & $F_{7}^6$&...& $F_{2^N-1}^1$&$F_{2^N-1}^2$&...&$F_{2^N-1}^{N!}$ \\ 
             \bottomrule
    	   \label{tab1}
    	\end{tabular}
    \end{center}
\end{table}

\subsubsection{The distance in Random Permutation Set theory}
Since RPST is a newly proposed theory, there is limited research on distance metrics between RPSs. The concept of ordered degree was proposed by Chen et al.\cite{chen2023distance} to quantify the similarity between any two permutation events. It was incorporated into the $\underline{D}$ matrix of the Jousselme distance\cite{jousselme2001new}, resulting in a distance measure between PMFs.

$\mathbf{Definition \ 2.7.}$ \textit{(Ordered degree of permutation events\cite{chen2023distance})} For any two permutation events $F_i^m$ and $F_j^n$ in $\mathcal{PES}(\mathcal{T})$, the ordered degree between $F_i^m$ and $F_j^n$ is defined as
\begin{equation}\label{eq12}
OD(F_i^m, F_j^n)=exp(-\frac{\sum_{\tau\in F_i^m\cap F_j^n}|rank_{F_i^m}(\tau)-rank_{F_j^n}(\tau)|}{|F_i^m\cup F_j^n|}).
\end{equation}
Where $rank_{F_i^m}(\tau)$ and $rank_{F_j^n}(\tau)$ indicate the position of the element $\tau$ in the permutations $F_i^m$ and $F_j^n$, respectively. For instance, if $F_i^m=(\tau_2\tau_1\tau_3)$ then $rank_{F_i^m}(\tau_1)=2$. Additionally, since $|F_i^m\overset{\leftarrow}{\cup} F_j^n|=|F_i^m\overset{\rightarrow}{\cup} F_j^n|=|F_i\cup F_j|$, Formula 
(\ref{eq12}) does not specifically display the left union or right union symbol.

$\mathbf{Definition \ 2.8.}$ \textit{(Chen et al.'s distance of RPS\cite{chen2023distance})} For two RPSs, $Perm_1$ and $Perm_2$, defined on the same FoD $\mathcal{T}$, the distance between them is defined by Chen et al. as
\begin{equation}\label{eq13}
d_{RPS}(Perm_1,Perm_2)=\sqrt{\frac{1}{2}(\mathbf{Perm_1}-\mathbf{Perm_2})\underline{RD}(\mathbf{Perm_1}-\mathbf{Perm_2})^T}
\end{equation}
Similarly, $\mathbf{Perm_1}$ and $\mathbf{Perm_2}$ are vectors corresponding to two RPSs, with dimensions $\Delta=\sum_{k=0}^{N}\frac{N!}{(N-k)!}$. The matrix $\underline{RD}$ is a $\Delta\times\Delta$ square matrix constructed based on the Jaccard index and the ordered degree. More specifically,
\begin{equation}\label{eq14}
\begin{aligned}
\underline{RD}(F_i^m,F_j^n)&=J(F_i^m,F_j^n)\times OD(F_i^m,F_j^n)\\
&=\frac{|F_i^m\cap F_j^n|}{|F_i^m\cup F_j^n|}\times exp(-\frac{\sum_{\tau\in F_i^m\cap F_j^n}|rank_{F_i^m}(\tau)-rank_{F_j^n}(\tau)|}{|F_i^m\cup F_j^n|})
\end{aligned}
\end{equation}

Alternatively, Formula (\ref{eq13}) can be expressed in another form:

\begin{equation}\label{eq15}
d_{RPS}(Perm_1,Perm_2)=\sqrt{\frac{1}{2}(||\mathbf{Perm_1}||^2+||\mathbf{Perm_2}||^2-2\langle\mathbf{Perm_1}, \mathbf{Perm_2}\rangle)}.
\end{equation}
Where $\langle\mathbf{Perm_1}, \mathbf{Perm_2}\rangle$ represents the scalar product of vectors $\mathbf{Perm_1}$ and $\mathbf{Perm_2}$, and is defined as
\begin{equation}\label{eq16}
\langle\mathbf{Perm_1}, \mathbf{Perm_2}\rangle=\sum_{i=0}^{\Delta-1}\sum_{j=0}^{\Delta-1}Perm_1(F_i^m)Perm_2(F_j^n)\times \underline{RD},
\end{equation}
with $F_i^m,F_j^n\in \mathcal{PES}(\mathcal{T})$ for $i, j=0, 1, ...,\Delta-1$. $||\mathbf{Perm}||^2$ represents the 2-norm of $\mathbf{Perm}$, given by $||\mathbf{Perm}||^2=\langle\mathbf{Perm}, \mathbf{Perm}\rangle$.

\subsection{An orness measure-based weight generation method}
The orness measure-based weight generation method was introduced by O’Hagan\cite{o1988aggregating} for the MEOWA operator. It determines the weights by maximizing entropy under the constraint of $orness$, which ensures the greatest uncertainty in the allocation of weights. 

$\mathbf{Definition \ 2.9.}$ \textit{(Ordered weighted average operator\cite{yager1988ordered})} Given an $n$-dimensional vector $\mathbf{a}=[a_1, a_2, ..., a_n]$, associate with an $n$-dimensional weight vector $\mathbf{w}=[\omega_1, \omega_2, ..., \omega_n]'$. The OWA operator is a mapping $f:[0, 1]^n\rightarrow[0,1]$, defined as
\begin{equation}\label{eq17}
f(a_1, a_2, ..., a_n)=\sum_{i=1}^{n}\omega_ia_{\rho(i)},
\end{equation}
where $\omega_i\in[0,1]$ and $\sum_{i=1}^{n}\omega_i=1$. Here $\rho$ is an index function, and $a_{\rho(i)}$ represents the $i$-th largest element in $\mathbf{a}$. The OWA offers a variety of aggregation operators depending on the weight settings. For instance, the weights $[1,0,...,0]'$, $[0,0,...,1]'$, and $[\frac{1}{n}, \frac{1}{n},..., \frac{1}{n}]'$ correspond to the $maximum$, $minimum$, and $mean$ aggregation operators, respectively.

$\mathbf{Definition \ 2.10.}$ \textit{(Orness measure\cite{yager1988ordered})} To better capture the performance of the OWA operator, Yager\cite{yager1988ordered} proposed the $orness$ measure as a characteristic measure, defined as follows:
\begin{equation}\label{eq18}
Orness(\mathbf{w})=\frac{1}{n-1}\sum_{i=1}^{n}(n-i)\omega_i.
\end{equation}
The $maximum$, $minimum$, and $mean$ aggregation operators correspond to cases where $Orness(\mathbf{w})$ takes values of $1$, $0$, and $0.5$, respectively. For simplicity, we will denote it as $Orn$ in the remainder of the article.

$\mathbf{Definition \ 2.11.}$ \textit{(The orness measure-based weight generation method\cite{o1988aggregating,filev1995analytic}}) The orness measure-based weight is generated by the maximum entropy model with the $Orn$ measure as a constraint, and its mathematical form is given by
\begin{equation}
MaxH(x)=-\sum_{i=1}^{n}\omega_iln\omega_i. \notag
\end{equation}
\begin{equation}\label{eq19}
s.t.\left\{ \begin{array}{l}
  Orn=\frac{\sum_{i=1}^{n}(n-i)\omega_i}{n-1};\\
  \sum_{i=1}^{n}\omega_i=1; \\
  0\leq \omega_i \leq 1.\\
\end{array} \right.
\end{equation}
For the constrained optimization problem (\ref{eq19}), Filev and Yager\cite{filev1995analytic} provided the corresponding analytical solution using the Lagrange multiplier method. The specific steps are as follows:

(1) Substitute the given $Orn$ value to construct an equation involving $h$, and then solve for the positive root of the equation.
\begin{equation}\label{eq20}
\sum_{i=1}^{n}(\frac{n-i}{n-1}-Orn)h^{n-i}=0
\end{equation}

(2) Calculate the weight vector $\mathbf{w}^{[Orn]}$ based on $h$.
\begin{equation}\label{eq21}
\omega_i^{[Orn]}=\frac{h^{n-i}}{\sum_{j=1}^nh^{n-j}}
\end{equation}

The important characteristic of the orness measure-based weight is that when $Orn\in [0,0.5)$, $\mathbf{w}^{[Orn]}$ is a non-decreasing sequence; when $Orn\in (0.5,1]$, $\mathbf{w}^{[Orn]}$ is a non-increasing sequence. Although this method was initially designed for generating the weights of the MEOWA aggregation operator, in this paper, we employ it solely to characterize the variation pattern of weights across different depths of permutations rather than to aggregate the Jaccard indices at different depths using the MEOWA operator.


\section{RFS and TBM Perspectives on RPS Distance Measures}\label{Different interpretations}
\begin{figure}[htbp!]
    \centering
    \includegraphics[width=1\textwidth]{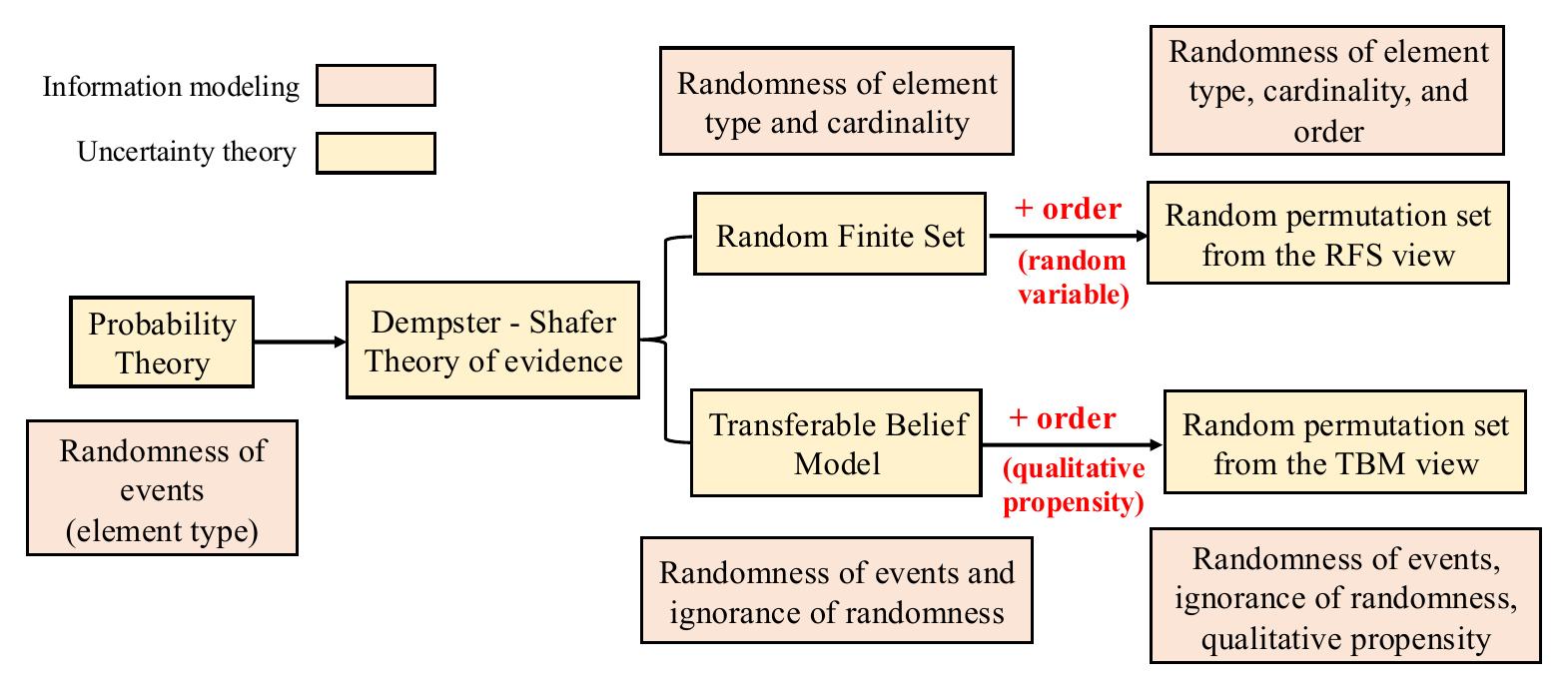}
    \caption{Different interpretations of RPS from RFS and TBM perspectives}
    \label{Interpretations}
\end{figure}

\begin{figure}[htbp!]
    \centering
    \includegraphics[width=0.6\textwidth]{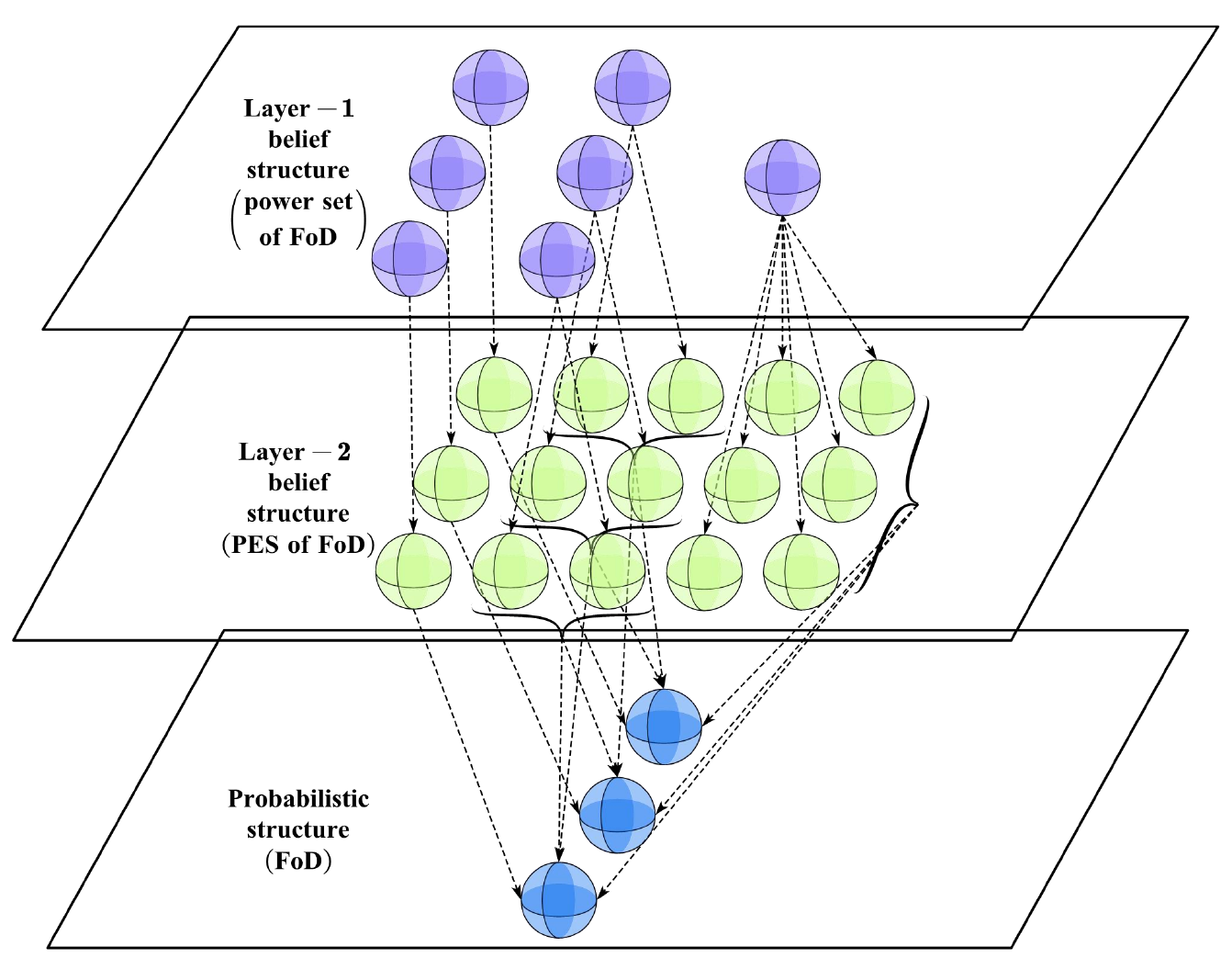}
    \caption{Layer-2 belief structure in a $3$-element FoD\cite{zhou2024conjunctive}}
    \label{rps-layer2}
\end{figure}

In this section, we discuss the different behaviors of RPS distance from the perspectives of RFS and TBM. Both RFS and TBM model uncertainty over the power set of the FoD and allow weights to be assigned to multi-element focal sets. Figure \ref{Interpretations} presents different interpretations of RPS from RFS and TBM perspectives. In RFS, a multi-element focal set results from the direct observation of two random variables: the cardinality and type of the elements. From the perspective of RFS, the newly introduced order in RPS serves as a third random variable added to the information distribution, alongside element type and element cardinality. PMF denotes the possibility or probability that the elements within an ordered proposition are the target elements. A typical application is in military threat assessment\cite{deng2022random}. To fully model an enemy invasion scenario, radar directly detects the permutation of enemy aircraft, capturing varying formations and appearance sequences, where each distinct permutation corresponds to a different event. From the perspective of epistemic uncertainty, the beliefs assigned to the multi-element focal set in TBM represent ignorance due to a lack of information. Zhou et al.\cite{zhou2024conjunctive} regarded RPST as a refinement of TBM. They proposed the RPS's layer-2 belief structure interpretation, where the order represents a qualitative propensity, indicating the decision-maker's belief transfer tendency. Under this interpretation, PMF quantifies the agent's degree of belief regarding the target elements within an ordered proposition. For $Perm(\tau_1\tau_2)=t$, it indicates that the belief $t$ held by the agent has a higher propensity to transfer to $\tau_1$. A layer-2 belief structure under a 3-element FoD is shown in Figure \ref{rps-layer2}. The layer-1 belief structure corresponds to the classic credal level, where beliefs are assigned to unordered focal sets. If no more information is currently available for fuse, beliefs are evenly distributed across singletons driven by the generalized insufficient reason principle. The layer-2 belief structure is a refined representation of layer-1, where these beliefs are further subdivided into corresponding ordered focal sets. At this point, the order is a symbolic constraint representing the propensity for belief transfer. It is weak qualitative information and insufficient to alter the focal set belief at the credal level. When no further information is available for fuse, this propensity manifests at the decision level.

Depending on different perspectives, the RPS distance measure exhibits distinct behavior and characteristics. Next, we present a simple example illustrating the different performances of the RPS distance measures under the above two perspectives.

$\mathbf{Example \ 3.1.}$ Consider three RPSs from distinct sources on the frame $\mathcal{T}=\{\tau_1, \tau_2, \tau_3\}$, with their corresponding PFMs as follows:
\begin{equation}
\begin{aligned}
&PMF_1(\mathcal{T}): Perm_1(F_3^1)=1\ \ \Rightarrow\ Perm_1((\tau_1\tau_2))=1, \\
&PMF_2(\mathcal{T}): Perm_2(F_7^1)=1\ \ \Rightarrow\ Perm_2((\tau_1\tau_2\tau_3))=1, \\
&PMF_3(\mathcal{T}): Perm_3(F_7^5)=1\ \ \Rightarrow\ Perm_3((\tau_3\tau_1\tau_2))=1. \notag
\end{aligned}
\end{equation}
From the perspective of RFS, $F_3^1$, $F_7^1$, and $F_7^5$ correspond to three distinct permutation events. $PMF_1$ indicates that two types of elements, $\tau_1$ and $\tau_2$, are observed in the order $\tau_1\tau_2$. In contrast, both $PMF_2$ and $PMF_3$ observe three types of elements $\tau_1$, $\tau_2$, and $\tau_3$, among which the common elements, $\tau_1$ and $\tau_2$, also appear in the order of $\tau_1\tau_2$. The distance between $Perm_1$ and $Perm_2$ arises from the difference in element cardinality and element type caused by the unobserved element $\tau_3$. The same holds for the distance between $Perm_1$ and $Perm_3$. Whether $\tau_3$ appears before or after $(\tau_1\tau_2)$, its position relative to $(\tau_1\tau_2)$ is equivalent. Therefore, under the interpretation of RFS, $d_{RPS}(Perm_1, Perm_2)=d_{RPS}(Perm_1, Perm_3)$. From the perspective of TBM, $PMF_1$ reflects the decision-maker's belief that the target element is either $\tau_1$ or $\tau_2$. The propensity $\tau_1\succ\tau_2$  indicates that, in the absence of further information for updating, and when a decision must be made, the decision-maker prefers $\tau_1$. Although both $PMF_2$ and $PMF_3$ point out the possible target elements are $\tau_1$, $\tau_2$, and $\tau_3$, they displayed different propensity. It can be readily observed that $PMF_1$ and $PMF_2$ share a similar belief transfer propensity $\tau_1\succ\tau_2$, whereas $PMF_3$ shows a preference for $\tau_3$. Compared to the distance between $Perm_1$ and $Perm_2$, which is solely attributed to the target element $\tau_3$, there additionally exists an inconsistency in qualitative propensity between $Perm_1$ and $Perm_3$. Therefore, under the interpretation of TBM, it follows that $d_{RPS}(Perm_1, Perm_2)<d_{RPS}(Perm_1, Perm_3)$.

This paper treats PMF as a refined representation of BPA, as illustrated by the layer-2 belief structure. The distance measure of RPS proposed in this paper applies only to information distributions in the form of layer-2 belief structure.

\section{Distance of Random Permutation Set under the TBM perspective}\label{proposed}
This section proposes an RPS distance measure method based on the cumulative Jaccard index following the TBM view, and its properties are analyzed from metric and structural aspects.

\subsection{Cumulative Jaccard index-based distance for RPS: definition and computation}
Given the characteristics of information distribution in the layer-2 belief structure, we first introduced a new concept of the cumulative Jaccard index to quantify the similarity between two permutations. Building on this, we incorporate it as a weighting matrix in the $L_2$ distance to formulate a distance measure for RPSs. The detailed calculation process is illustrated through two numerical examples.

\begin{figure}[htbp!]
    \centering
    \includegraphics[width=0.8\textwidth]{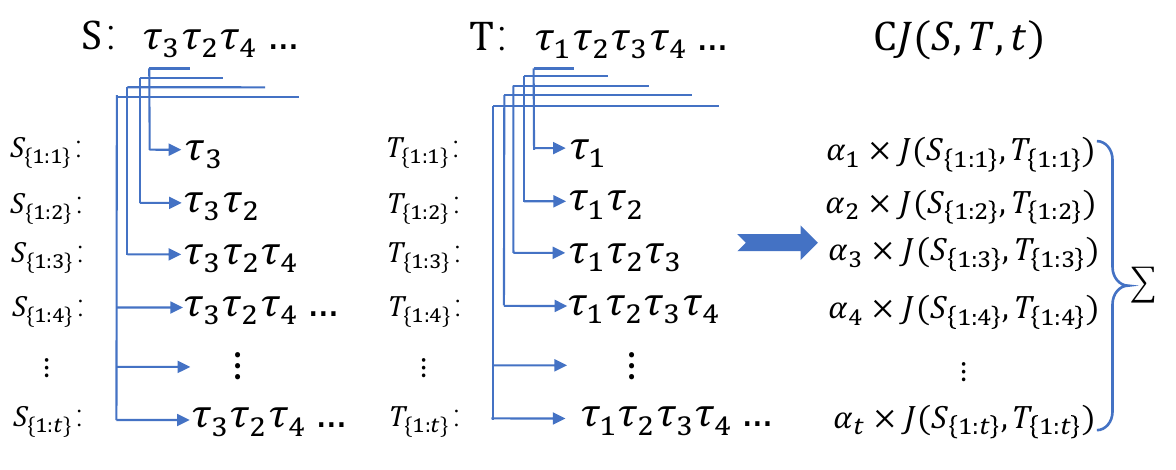}
    \caption{Example diagram of cumulative Jaccard index computation}
    \label{Cumulative Jaccard index}
\end{figure}

$\mathbf{Definition \ 4.1.}$ \textit{(Cumulative Jaccard index)} Given two permutations $S$ and $T$, where $S_i$ represents the $i$-th element in permutation $S$. The notation $S_{\{l_1:l_2\}} = \{S_i: l_1\le i\le l_2\}$ defines a sub-permutation of $S$, consisting of elements from position $l_1$ to $l_2$. The cumulative Jaccard index between permutations $S$ and $T$ at depth $t$ is calculated as follows:
\begin{equation}\label{eq22}
\begin{aligned}
CJ(S, T, t)&=\sum_{d=1}^{t}\alpha_d\times J(S_{\{1:d\}}, T_{\{1:d\}}) \\
&=\sum_{d=1}^{t}\alpha_d\times \frac{|S_{\{1:d\}}\cap T_{\{1:d\}}|}{|S_{\{1:d\}}\cup T_{\{1:d\}}|}.
\end{aligned}
\end{equation}
Where $1 \le t \le max(|S|, |T|)$. If not specified, $t$ defaults to $max(|S|, |T|)$. In particular, $t$ is allowed to take any integer value between $1$ and $ max(|S|, |T|)$, which means that the decision-maker only focuses on the first $t$ target elements and discards redundancy. $J(S_{\{1:d\}}, T_{\{1:d\}})$ is the Jaccard index of permutations $S$ and $T$ at depth $d$. $\alpha_d$ is the weight at depth $d$, which can be subjectively assigned or objectively generated according to the actual application scenario. To enhance clarity, the detailed calculation process is illustrated in Figure \ref{Cumulative Jaccard index}.

$\mathbf{Definition \ 4.2.}$ \textit{(Cumulative Jaccard index with orness measure-based weights)} In this paper, we focus on the orness measure-based weights introduced in Definition 2.11. The weights in $CJ(S, T, t)$ are set to $\alpha_d=\omega_d^{[Orn]}$, and then

\begin{equation}\label{eq23}
\begin{aligned}
CJ^{[Orn]}(S, T, t)&=\sum_{d=1}^{t}\omega_d^{[Orn]}\times J(S_{\{1:d\}}, T_{\{1:d\}}) \\
&=\sum_{d=1}^{t}\omega_d^{[Orn]}\times \frac{|S_{\{1:d\}}\cap T_{\{1:d\}}|}{|S_{\{1:d\}}\cup T_{\{1:d\}}|}.
\end{aligned}
\end{equation}
 According to the definition, the value of $CJ^{[Orn]}$ falls in $[0, 1]$. The extreme value of $1$ indicates two identical permutations, while the extreme value of $0$ signifies completely different permutations. 
 
 The cumulative Jaccard index quantifies the similarity between two permutations by calculating the cumulative sum of the Jaccard index of the corresponding sub-permutations at each depth. The index naturally possesses a top-weightiness property, which reflects the order factor in the permutation by assigning priority and repeated emphasis to the top-ranked elements. $Orn\in [0,1]$ is provided as a parameter and can be used to adjust the weights at different depths. Different parameter settings reflect different preferences of decision-makers. When $Orn=1$, there is only the top weight, which is $1$, indicating that only the frontmost element of the permutation is considered. When $Orn \in (0.5, 1]$, the weight gradually decreases from the top to the bottom, while the top weight increases with rising values of $Orn$, further enhancing the top-weightiness property. On the contrary, when $Orn \in [0, 0.5)$, the weight gradually increases from the top to the bottom, with the bottom weight rising as $Orn$ decreases, thereby continuously weakening the top-weightiness property. Until $Orn=0$, only the bottom weight is $1$, reflecting a decision-maker’s preference for permutations without the top-weightiness property. At this time, $CJ^{[Orn]}$ degenerates into the classic Jaccard index. The choice of parameters $Orn$ and $t$ should always be guided by practical considerations relevant to the specific application. If no additional declaration is made, the default value of $Orn$ is $0.5$. At this time, the decision-maker adopts a neutral stance, with weights uniformly distributed across all depths.

$\mathbf{Example \ 4.1.}$ Consider two permutations, $S: \tau_3\tau_2\tau_4$ and $T: \tau_1\tau_2\tau_3\tau_4$. Table \ref{tab2} presents the Jaccard index for the sub-permutations of $S$ and $T$ at each depth.
 \begin{table}[!htbp]
    	\begin{center}
        	\caption{Jaccard index of sub-permutations of $S$ and $T$ at different depths}
    		\begin{tabular}{cllccc}
    			\toprule
    			$Depth$ &\ \ \ \ $S$ &\ \ \ \ \  $T$ &$Jaccard\ index$ \\
    			\midrule
                    \specialrule{0em}{1pt}{1pt}
    			1 & $\tau_3$ & $\tau_1$ & $0$	\\
    			\specialrule{0em}{1pt}{1pt}
    			2 & $\tau_3,\tau_2$ &$\tau_1,\tau_2$ &$1/3$ \\
        		\specialrule{0em}{1pt}{1pt}
    			3 &$\tau_3,\tau_2,\tau_4$  &$\tau_1,\tau_2,\tau_3$&$1/2$  \\
        		\specialrule{0em}{1pt}{1pt}
    			4  &$\tau_3,\tau_2,\tau_4$  &$\tau_1,\tau_2,\tau_3,\tau_4$ &$3/4$\\
    			\bottomrule
    			\label{tab2}
    		\end{tabular}
    	\end{center}
    \end{table}
When $Orn=0.7$, the weight vector calculated by formulas (\ref{eq20}) and (\ref{eq21}) is $\mathbf{w}^{[0.7]}=[0.4614, 0.2756, 0.1647, 0.0984]'$. The cumulative Jaccard index of permutations $S$ and $T$ is calculated as
\begin{equation}
\begin{aligned}
CJ^{[0.7]}(S, T)&=0.4614\times0+0.2756\times\frac{1}{3}+0.1647\times\frac{1}{2}+0.0984\times\frac{3}{4} \\ 
&=0.2480.  \notag
\end{aligned}
\end{equation}

Similarly, the $CJ^{[Orn]}$ values of $S$ and $T$ when $Orn=0, 0.1, 0.2, ..., 1$ are listed in Table \ref{tab3}. 
 \begin{table}[!htbp]\small
    	\begin{center}
        	\caption{The cumulative Jaccard index of $S$ and $T$ when $Orn$ takes different values}
    		\begin{tabular}{ccccccc}
    			\toprule
    			         & $Orn=0.0$ &$Orn=0.1$ &$Orn=0.2$&$Orn=0.3$ &$Orn=0.4$  & $Orn=0.5$ \\
    			\midrule
                    \specialrule{0em}{1pt}{1pt}
    			$CJ^{[Orn]}(S, T)$ &0.7500  &0.6786  &0.6089 & 0.5387& 0.4678 & 0.3958 	\\
           			\toprule
    			    & $Orn=0.5$ &$Orn=0.6$ &$Orn=0.7$&$Orn=0.8$ &$Orn=0.9$ &$Orn=1.0$  \\
    			\midrule
                    \specialrule{0em}{1pt}{1pt}
    			$CJ^{[Orn]}(S, T)$ & 0.3958 & 0.3227& 0.2480& 0.1701 &0.0902 &0.0000	\\
    			\bottomrule
    			\label{tab3}
    		\end{tabular}
    	\end{center}
\end{table}

It can be observed that $S$ and $T$ begin to have common elements after the depth reaches 2. As the depth increases, the number of overlapping elements increases, making the Jaccard index of $S$ and $T$ increase with increasing depth. In this case, a more significant top weight results in a smaller $CJ^{[Orn]}$ value, while a more substantial bottom weight leads to a larger $CJ^{[Orn]}$ value. When $0 \leq Orn \leq 0.5$, the bottom weight increases as $Orn$ decreases. Consequently, the smaller the $Orn$, the larger the value of $CJ^{[Orn]}$. When $Orn = 0$, the bottom weight reaches 1, and at this point, $CJ^{[Orn]}$ equals the Jaccard index of $S$ and $T$ at depth 4, which is $\frac{3}{4}$. When $0.5 \leq Orn \leq 1$, the top weight increases with an increase in $Orn$, leading to a corresponding decrease in the $CJ^{[Orn]}$ value. At $Orn = 1$, the top weight attains its maximum value of 1, resulting in the decision-maker's exclusive focus on the first element of the permutation, which consequently yields a $CJ^{[Orn]}$ value of 0. It is worth mentioning that the selection of $Orn$ does not alter the intrinsic nature of the property of top-weightiness; instead, it can only enhance or weaken this property. In addition, the cumulative Jaccard index can perform arbitrary depth truncation on the permutation by specifying $t$. For example, when $Orn = 0.5$, the cumulative Jaccard index of $S$ and $T$ at depths 1 to 4 is 0, 0.1667, 0.2778, and 0.3958, respectively. The arbitrary truncation property empowers decision-makers to selectively focus on the elements they consider most pertinent. 

$\mathbf{Definition \ 4.3.}$ \textit{(The proposed distance measure between RPSs)} Suppose there are two RPSs, $Perm_1$ and $Perm_2$, defined on the same FoD $\mathcal{T}$, the proposed distance measure between $Perm_1$ and $Perm_2$ is defined as
\begin{equation}\label{eq24}
d_{RPS}^{[Orn]}(Perm_1, Perm_2)=\sqrt{\frac{1}{2}(\mathbf{Perm_1}-\mathbf{Perm_2})\underline{CD}^{[Orn]}(\mathbf{Perm_1}-\mathbf{Perm_2})^T}
\end{equation}
Where $\mathbf{Perm_1}$ and $\mathbf{Perm_2}$ are vectors corresponding to two RPSs, with dimensions $\Delta=\sum_{k=0}^{N}\frac{N!}{(N-k)!}$. The normalization factor $\frac{1}{2}$ is used to guarantee that $0\le d_{RPS}^{[Orn]}\le 1$. The weighting matrix $\underline{CD}^{[Orn]}$ is a $\Delta\times\Delta$ cumulative Jaccard index matrix. More specifically,
\begin{equation}\label{eq25}
\begin{aligned}
\underline{CD}^{[Orn]}(F_i^m, F_j^n, t)&=CJ^{[Orn]}(F_i^m,F_j^n, t)\\
&=\sum_{d=1}^{t}\omega_d^{[Orn]}\times \frac{|{F_i^m}_{\{1:d\}}\cap {F_j^n}_{\{1:d\}}|}{|{F_i^m}_{\{1:d\}}\cup {F_j^n}_{\{1:d\}}|}
\end{aligned}
\end{equation}
where $1 \le t \le max(|F_i^m|, |F_j^n|)$, $Orn\in [0,1]$. In the absence of explicit specification, the default parameters are set to $t=max(|F_i^m|, |F_j^n|)$ and $Orn=0.5$.

Alternatively, Formula (\ref{eq24}) can be expressed in another form:

\begin{equation}\label{eq26}
d_{RPS}^{[Orn]}(Perm_1, Perm_2)=\sqrt{\frac{1}{2}||\mathbf{Perm_1}||^2+||\mathbf{Perm_2}||^2-2\langle\mathbf{Perm_1}, \mathbf{Perm_2}\rangle}.
\end{equation}
Where $\langle\mathbf{Perm_1}, \mathbf{Perm_2}\rangle$ represents the scalar product of vectors $\mathbf{Perm_1}$ and $\mathbf{Perm_2}$, and is defined as
\begin{equation}\label{eq27}
\langle\mathbf{Perm_1}, \mathbf{Perm_2}\rangle=\sum_{i=0}^{\Delta-1}\sum_{j=0}^{\Delta-1}Perm_1(F_i^m)Perm_2(F_j^n)\times \underline{CD}^{[Orn]}(F_i^m, F_j^n, t),
\end{equation}
with $F_i^m,F_j^n\in \mathcal{PES}(\mathcal{T})$ for $i, j=0, 1, ...,\Delta-1$. $||\mathbf{Perm}||^2$ represents the 2-norm of $\mathbf{Perm}$, given by $||\mathbf{Perm}||^2=\langle\mathbf{Perm}, \mathbf{Perm}\rangle$.

$\mathbf{Example \ 4.2.}$  Consider two RPSs defined on $\mathcal{T}=\{\tau_1, \tau_2, \tau_3\}$ and their PMFs are as follows:
\begin{equation}
\begin{aligned}
&PMF_1(\mathcal{T}): Perm_1(F_{1}^1)=0.4, \ \ \ \ Perm_1(F_{5}^1)=0.3, \ \ \ \ Perm_1(F_{5}^2)=0.3; \\
&\ \ \ \ \ \ \Rightarrow \ \ \ \ \  Perm_1((\tau_1))=0.4,\ \ Perm_1((\tau_1\tau_3))=0.3,\ \ Perm_1(\tau_3\tau_1)=0.3.\\
&PMF_2(\mathcal{T}): Perm_2(F_{2}^1)=0.4, \ \ \ \ Perm_2(F_{5}^1)=0.1, \ \ \ \ Perm_2(F_{7}^1)=0.15, \\
&\ \ \ \ \ \ \ \ \ \ \ \ \ \ \ \  Perm_2(F_{7}^4)=0.35;   \\
&\ \ \ \ \ \ \Rightarrow \ \ \ \ \  Perm_2((\tau_2))=0.4,\ \ Perm_2((\tau_1\tau_3))=0.1,\ \ Perm_2((\tau_1\tau_2\tau_3))=0.15,\\
& \ \ \ \ \ \ \  \ \ \ \ \ \ \ \ \  Perm_2((\tau_2\tau_3\tau_1))=0.35.  \notag\\
\end{aligned}
\end{equation}

The vector representations of $RPS_1$ and $RPS_2$ are denoted as $\mathbf{Perm_1}$ and $\mathbf{Perm_1}$, respectively.

$\mathbf{Perm_1}=[0.4, 0, 0, 0, 0, 0.3, 0.3, 0, 0, 0, 0, 0, 0, 0, 0];$

$\mathbf{Perm_2}=[0, 0.4, 0, 0, 0, 0.1, 0, 0, 0, 0.15, 0, 0, 0.35, 0, 0].$

The cumulative Jaccard index matrix $\underline{CD}^{[0.5]}$ is constructed using the formula (\ref{eq25}). For the permutation events $F_{5}^1: (\tau_1\tau_3)$ and $F_{7}^1: (\tau_1\tau_2\tau_3)$, the specific calculation steps of $\underline{CD}^{[0.5]}(F_{5}^1, F_{7}^1)$ are as follows.

\begin{equation}
\begin{aligned}
\underline{CD}^{[0.5]}(F_{5}^1, F_{7}^1)&=CJ^{[0.5]}(F_{5}^1, F_{7}^1)\\
&=\sum_{d=1}^{3}\omega_d^{[0.5]}\times \frac{|{F_{5}^1}_{\{1:d\}}\cap {F_{7}^1}_{\{1:d\}}|}{|{F_{5}^1}_{\{1:d\}}\cup {F_{7}^1}_{\{1:d\}}|}  \\
&=\frac{1}{3}\times1+\frac{1}{3}\times\frac{1}{3}+\frac{1}{3}\times\frac{2}{3} \\
&=0.6667   \notag
\end{aligned}
\end{equation}

\begin{figure}[htbp!]
    \centering
    \includegraphics[width=0.65\textwidth]{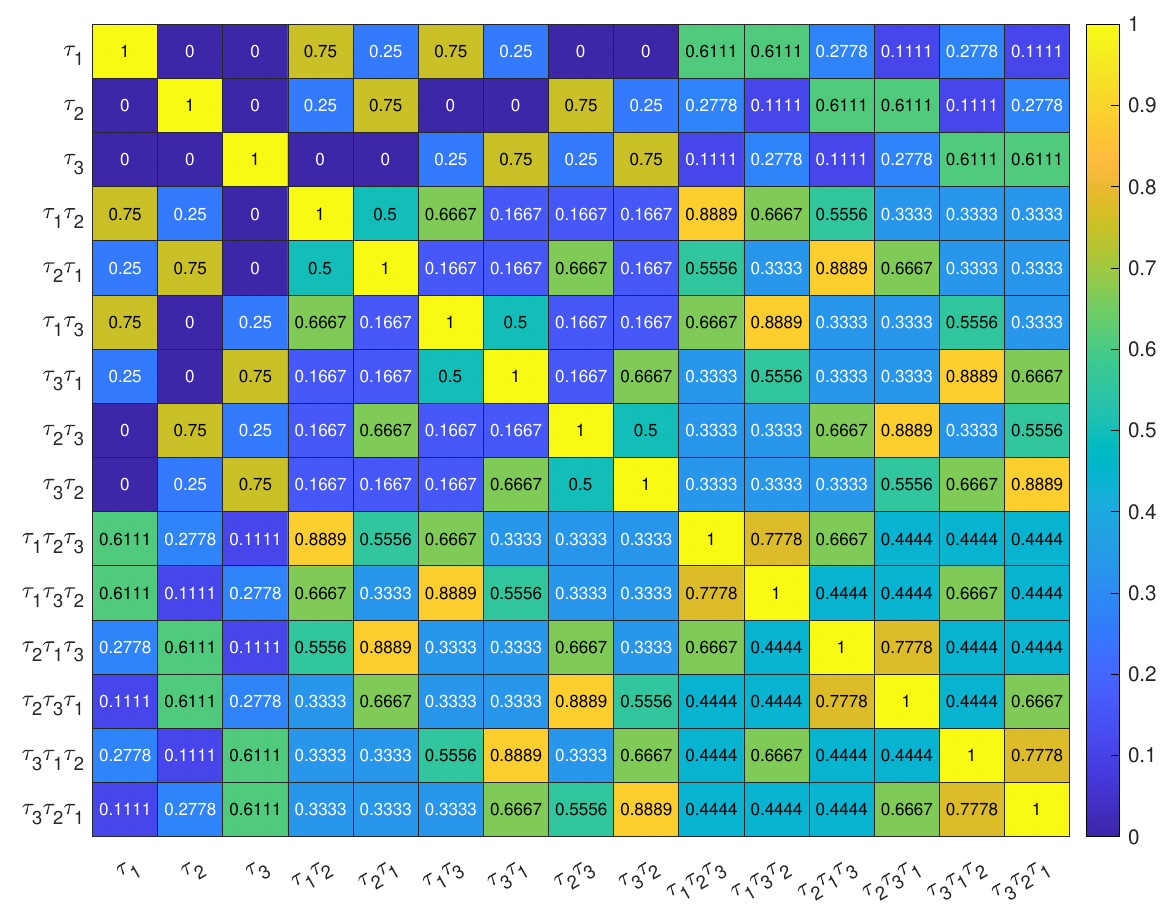}
    \caption{Heatmap of $\underline{CD}^{[0.5]}$ in a $3$-element FoD}
    \label{Heatmap0}
\end{figure}

Similarly, the cumulative Jaccard Index for all possible permutation events is calculated. The heatmap of the $\underline{CD}^{[0.5]}$ is presented in Figure \ref{Heatmap0}.

Finally, the distance between $RPS_1$ and $RPS_2$ is calculated using Formula (\ref{eq24}) as follows:
\begin{equation}
\begin{aligned}
d_{RPS}^{[0.5]}(Perm_1, Perm_2)&=\sqrt{\frac{1}{2}(\mathbf{Perm_1}-\mathbf{Perm_2})\underline{CD}^{[0.5]}(\mathbf{Perm_1}-\mathbf{Perm_2})^T}\\
&=0.6305   \notag
\end{aligned}
\end{equation}

\subsection{Properties of the proposed RPS distance measure}
Next, we examine the properties of the proposed RPS distance measure from metric and structural aspects.

\subsubsection{Metric properties}
As mentioned in Definition 2.1, this paper mainly considers the following metric properties.

 $(d1)$ \textit{Nonnegativity: $d_{RPS}^{[Orn]}(Perm_1, Perm_2)\ge 0$.}

$(d2)$ \textit{Symmetry: $d_{RPS}^{[Orn]}(Perm_1, Perm_2) = d_{RPS}^{[Orn]}(Perm_2, Perm_1)$.}

$(d3)$ \textit{Definiteness: $d_{RPS}^{[Orn]}(Perm_1, Perm_2) = 0 \Leftrightarrow  Perm_1 = Perm_2$}.

$(d4)$ \textit{Triangle inequality: $d_{RPS}^{[Orn]}(Perm_1, Perm_2) \leq d_{RPS}^{[Orn]}(Perm_1, Perm_3) + d_{RPS}^{[Orn]}(Perm_2, Perm_3), \forall Perm_3$.} \\
Among them, property $(d3)$ is split into the following two separate properties:

\ \ \ \ \ \ $(d3)'$ \textit{Reflexivity: $d_{RPS}^{[Orn]}(Perm_1, Perm_1) = 0$}.

\ \ \ \ \ \ $(d3)''$ \textit{Separability: $d_{RPS}^{[Orn]}(Perm_1, Perm_2) = 0 \Rightarrow  Perm_1 = Perm_2$}.

Properties $(d2)$ and $(d3)'$ are easy to prove, while properties $(d1)$, $(d3)''$, and $(d4)$ are related to the positive definiteness of the weighting matrix. In DST, the distance of Jousselme et al.\cite{jousselme2001new} is a full metric since its weighting matrix, the Jaccard index matrix, has been proven to be positive definite for any size of FoD\cite{bouchard2013proof}. For RPST, the positive definiteness of the weighting matrix set in the RPS distance measure is also a prerequisite to guarantee the metric properties of the RPS distance measure. This has also become one of the considerations when we formulate the distance between the permutation set information distributions.

Table \ref{tab4} lists the minimum eigenvalues $\lambda_{min}$ of the classic Jaccard index matrix $\underline{D}$ used in Jousselme et al.' distance\cite{jousselme2001new}, the weighting matrix $\underline{RD}$ in the RPS distance measure proposed by Chen et al.\cite{chen2023distance}, and the proposed cumulative Jaccard matrix $\underline{CD}$ under the PES structure. Here, we only show the results when the FoD size ranges from 1 to 7, which is enough to illustrate the problem. 

 \begin{table}[!htbp]
    	\begin{center}
    		\setlength{\tabcolsep}{0.5cm}
    		\caption{The minimum eigenvalue $\lambda_{min}$ of weighting matrix in RPS distance measure for different FoD sizes}
    		\begin{tabular}{cccc}
    			\toprule
                     $Size\ of\ FoD$&  \multicolumn{3}{c}{$\lambda_{min}$}  \\
                    \cmidrule(lr){2-4}
             			  &$\underline{D}$& $\underline{RD}$ &$\underline{CD}$   \\
                   &Jousselme et al.\cite{jousselme2001new}& Chen et al.\cite{chen2023distance} &Proposed method   \\
    			\midrule
    			\specialrule{0em}{1pt}{1pt}
    			 $|\mathcal{T}|=1$ &1.0000 &1.0000  &1.0000 \\
        		\specialrule{0em}{1pt}{1pt}
    			 $|\mathcal{T}|=2$  &$0.0000$  &0.3599 &0.1910   \\
        		\specialrule{0em}{1pt}{1pt}
    			 $|\mathcal{T}|=3$   &$0.0000$ &0.1763 &-0.0167 \\
        		\specialrule{0em}{1pt}{1pt}
    			 $|\mathcal{T}|=4$ & $0.0000$ &-0.0172 &-0.4540  \\
                    \specialrule{0em}{1pt}{1pt}
    		   $|\mathcal{T}|=5$ &$0.0000$ &-0.7462 &-2.8544 \\
                    \specialrule{0em}{1pt}{1pt}
    		   $|\mathcal{T}|=6$ & $0.0000$ &-5.5261  &-18.4412 \\
                    \specialrule{0em}{1pt}{1pt}
    		   $|\mathcal{T}|=7$ & $0.0000$ &-41.9742  &-133.1948 \\
    			\bottomrule
    			\label{tab4}
    		\end{tabular}
    	\end{center}
    \end{table}

It can be observed that the Jaccard index matrix $\underline{D}$ has zero eigenvalues when the size of the FoD exceeds 2, indicating that it is actually semi-positive definite under the permutation set structure. This is because different random permutation sets can be mapped into the same random set when ordered focal sets are transformed into ordinary sets by forgetting the order of their elements. When the size of FoD is $|\mathcal{T}|=3$, the matrix $\underline{CD}$ begins to exhibit negative eigenvalues. As $|\mathcal{T}|$ increases exceeding 3, all matrices are no longer positive definite. Consequently, any RPS distance measure method that employs these matrices as weighting matrices fails to satisfy the properties $(d1)$, $(d3)''$, and $(d4)$. We attribute it to the structural characteristics of the permutation set. Similarity measures commonly used for sets or permutations, including those that ensure positive definiteness under the power-set structure, often struggle to maintain positive definiteness within the permutation-set framework. Although this may have a limited impact in certain practical scenarios, we still provide a solution to modify the matrix to ensure full metric compliance.

$\mathbf{Definition \ 4.4.}$ \textit{(Corrected cumulative Jaccard index matrix)} When the minimum eigenvalue $\lambda_{min}$ of the cumulative Jaccard index matrix $\underline{CD}^{[Orn]}$ is non-positive, the non-positive eigenvalues are adjusted to small positive values to make the matrix positive definite. The resulting corrected cumulative Jaccard index matrix is defined as
\begin{equation}\label{eq28}
    \underline{CD}^{[Orn, \lambda_{min}]} =\begin{cases}
          \underline{CD}^{[Orn]}  & \lambda_{min}>0, \\
          \frac{1}{|\lambda_{min}|+\epsilon} (\underline{CD}^{[Orn]}+(|\lambda_{min}|+\epsilon)I)   & \lambda_{min}\le 0.
    \end{cases}
\end{equation}
Where $I$ is the identity matrix and $\epsilon$ is the correction parameter used to ensure the corrected eigenvalue $\lambda_{min}'>0$. $\epsilon$ can be an infinitely small positive number. In this paper, we take $\epsilon=10^{-12}$.

Naturally, the corresponding distance between RPSs is written as
\begin{equation}\label{eq29}
d_{RPS}^{[Orn, \lambda_{min}]}(Perm_1, Perm_2)=\sqrt{\frac{1}{2}(\mathbf{Perm_1}-\mathbf{Perm_2})\underline{CD}^{[Orn, \lambda_{min}]}(\mathbf{Perm_1}-\mathbf{Perm_2})^T}.
\end{equation}

\begin{figure}[htbp!]
    \centering
    \includegraphics[width=0.65\textwidth]{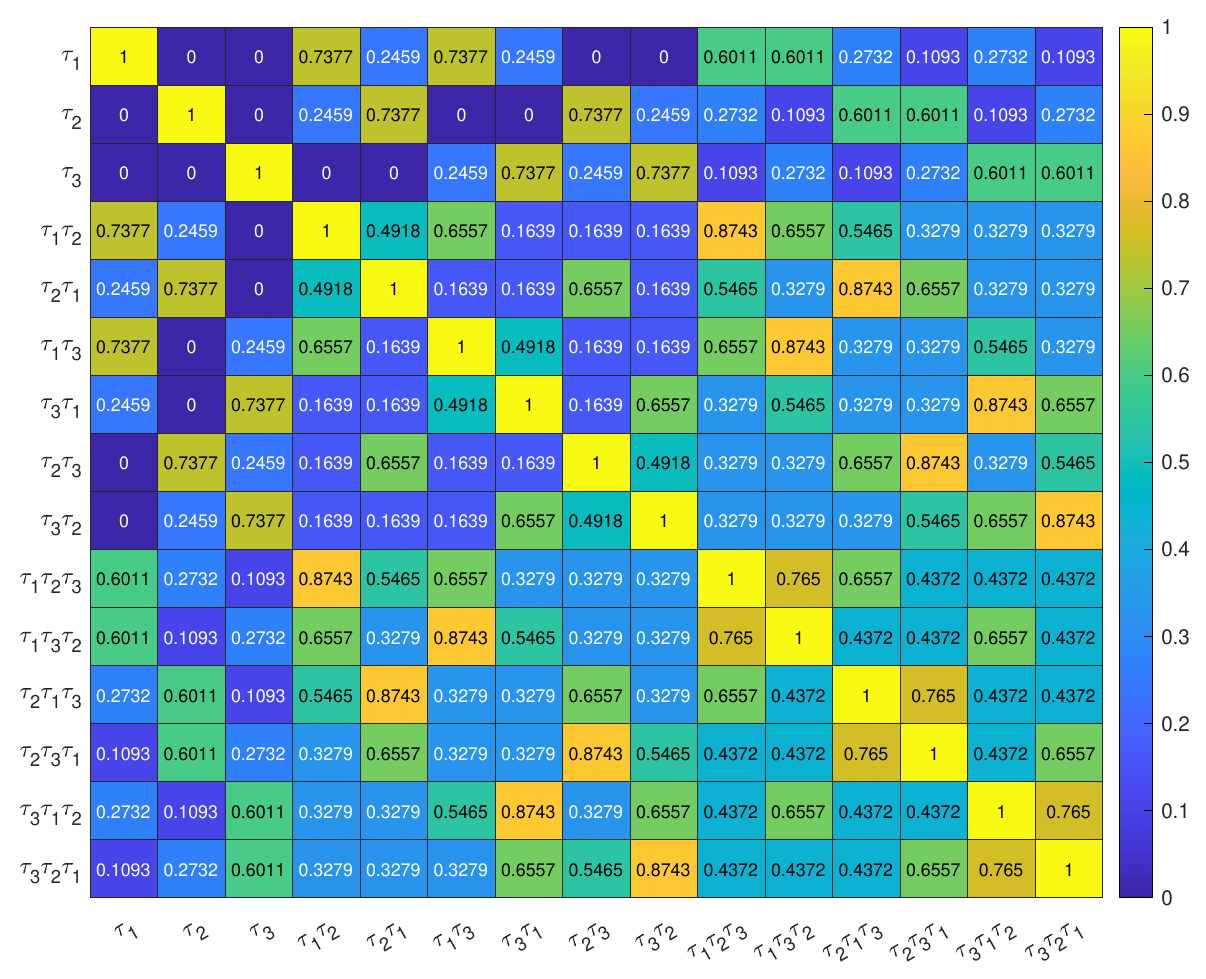}
    \caption{Heatmap of $\underline{CD}^{[0.5, -0.0167]}$ in a $3$-element FoD}
    \label{Heatmap}
\end{figure}

Let us revisit the two RPSs in Example 4.2, which are defined on a 3-element FoD. At this time, the cumulative Jaccard matrix $\underline{CD}^{[0.5]}$ actually has a minimum eigenvalue $\lambda_{min}=-0.0167$. Use the formula (\ref{eq28}) to correct it, and the corrected cumulative Jaccard index matrix $\underline{CD}^{[0.5, -0.0167]}$ is shown in Figure \ref{Heatmap}. The distance between the RPSs calculated using the corrected weighting matrix is $d_{RPS}^{[0.5,-0.0167]}(Perm_1, Perm_2)=0.6292$.

As the size $N$ of FoD continues to increase, the size of the PES $|\mathcal{PES}(\mathcal{T})|$ exhibits factorial growth.\footnote{$\lim_{N\rightarrow\infty}|\mathcal{PES}(\mathcal{T})|\approx e\cdot N!$} Consequently, the dimension of the weighting matrix keeps increasing, resulting in an explosive escalation in the computational complexity of calculating distances between RPSs. Nevertheless, it is not difficult to find that in practical applications, the number of focal sets in common RPS is not significant $(|\mathcal{OF}|\ll|\mathcal{PES}(\mathcal{T})|)$; that is, $\mathbf{Perm}$ is a sparse vector in most cases. Actually, when we calculate the distance between RPSs, we usually ignore the non-focal sets to simplify the calculation. From the perspective of engineering practice, this is also a better choice. 

We still use Example 4.2 as an illustration. Considering only the ordered focal sets in $\mathcal{OF}_1$ and $\mathcal{OF}_2$, $\mathbf{Perm_1}$ and $\mathbf{Perm_2}$ are written as

$\mathbf{Perm_1}=[0.4, 0, 0.3, 0.3, 0, 0];$

\begin{figure}[htbp!]
    \centering
    \includegraphics[width=0.4\textwidth]{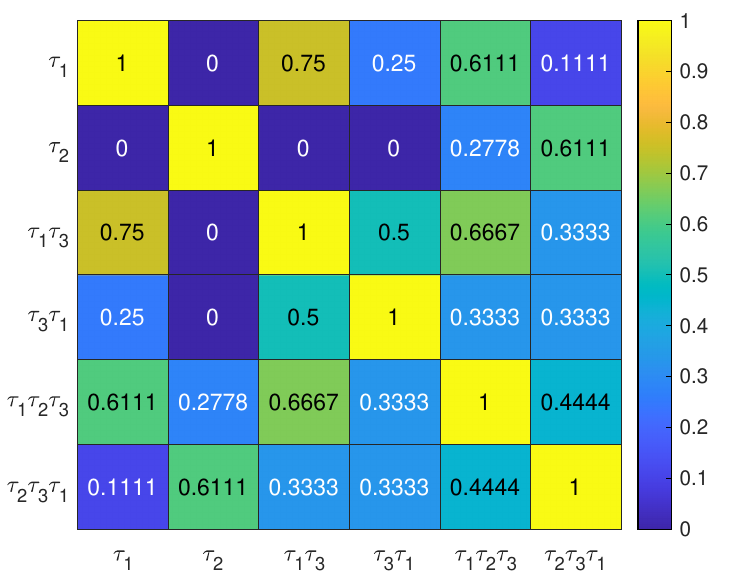}
    \caption{Heatmap of $\underline{CD}^{[0.5]}$ after omitting the non-focal set in Example 4.2}
    \label{Heatmap2}
\end{figure}

$\mathbf{Perm_2}=[0, 0.4, 0.1, 0, 0.15, 0.35].$ \\
The corresponding cumulative Jaccard matrix $\underline{CD}^{[0.5]}$ is shown in Figure \ref{Heatmap2}. Since $\underline{CD}^{[0.5]}$ becomes a positive definite matrix after removing the non-focal sets, no further correction is required. The distance between RPSs is thus computed as $d_{RPS}^{[0.5]}(Perm_1, Perm_2)=0.6305$. It is worth noting that the removal of unused non-focal sets does not affect the final results. In fact, we observed that, in most cases, the weighting matrix obtained after omitting the non-focal sets is positive definite. The distance values between RPSs presented in the subsequent numerical examples in this paper are all calculated based on representations with non-focal sets excluded. The specific calculation process is shown in Algorithm \ref{algorithm}.

\begin{algorithm}[htbp!] \SetKwInOut{Input}{Input}\SetKwInOut{Output}{Output}
	\Input{Two RPSs: $Perm_1$, $Perm_2$; Parameter: $Orn$, $t$.} 
	\Output{The distance value between $RPS_1$ and $RPS_2$: $d_{RPS}^{[Orn, \lambda_{min}]}(Perm_1, Perm_2)$. }
	 \BlankLine 
       \% Represent input RPSs in vector form (containing only the focal sets in $\mathcal{OF}_1$ and $\mathcal{OF}_2$)\;
      $\mathbf{Perm_1}, \mathbf{Perm_2}\leftarrow Perm_1, Perm_2$\;
       \% Construct the cumulative Jaccard index matrix\;
       $s\leftarrow$ length of vector $\mathbf{Perm_1}$ \;
       Initialize $\underline{CD}^{[Orn]}$ as an $s \times s$ identity matrix $I_s$\;
      \For{$i\leftarrow 1$ \KwTo $s-1$}{
      \For{$j\leftarrow i+1$ \KwTo $s$ }{
      $temp_A\leftarrow$ the focal set corresponding to the vector $\mathbf{Perm_1}[i]$ \;
      $temp_B\leftarrow$ the focal set corresponding to the vector $\mathbf{Perm_1}[j]$ \;
      \If{$temp_A\cap temp_B==\emptyset$}
      {
      $\underline{CD}^{[Orn]}[i, j]\leftarrow 0$\;
      }
      \Else{$\underline{CD}^{[Orn]}[i, j]\leftarrow$ based on parameters $Orn$ and $t$, calculate cumulative Jaccard index of $temp_A$ and $temp_B$ by Eq. (\ref{eq25}), where the weight vector $\mathbf{w}^{[Orn]}$ is obtained by Eq. (\ref{eq20}) to Eq. (\ref{eq21}); }
      }
      }
      $\underline{CD}^{[Orn]}\leftarrow\underline{CD}^{[Orn]}+\underline{CD}^{T[Orn]}-I_s$ \;
      \% Check if the matrix $\underline{CD}^{[Orn]}$ is positive definite \;
      $Eigenvalues \leftarrow eig(\underline{CD}^{[Orn]})$\;
      $\lambda_{min}\leftarrow min(eig(\underline{CD}^{[Orn]}))$\;
      \If{$\lambda_{min}\leq 0$}{
      $\underline{CD}^{[Orn, \lambda_{min}]} \leftarrow$ correct the matrix $\underline{CD}^{[Orn]}$ using Eq. (\ref{eq28})\;}
      \Else{ $\underline{CD}^{[Orn, \lambda_{min}]} \leftarrow \underline{CD}^{[Orn]} $}
      \% Calculate the distance value between $RPS_1$ and $RPS_2$ \;
      $d_{RPS}^{[Orn, \lambda_{min}]}(Perm_1, Perm_2)\leftarrow$ calculate the distance value by Eq. (\ref{eq29})  \;
      \Return $d_{RPS}^{[Orn, \lambda_{min}]}(Perm_1, Perm_2)$ \;
     \caption{Cumulative Jaccard index-based RPS distance algorithm} \label{algorithm}
\end{algorithm}

\subsubsection{Structural property of permutation mass functions}
Besides the above-mentioned metric properties, a distance between permutation mass functions should satisfy some structural properties particular to the nature of permutation mass functions. On the layer-1 belief structure, Jousselme and Maupin\cite{jousselme2012distances} established three fundamental structural properties, including \textit{strong structural property}, \textit{weak structural property}, and \textit{structural dissimilarity}, that should be considered when defining a distance measure for belief functions. Analogously, for the distance measure between information distributions in the form of layer-2 belief structure, we specially formulate the following properties:

$(sp1)$ Strong structural property (\textit{interaction between ordered focal elements}): 
A distance measure $d$ between two permutation mass functions $Perm_1$ and $Perm_2$ is considered \textit{strongly structural} if its formulation incorporates the interactions between the ordered focal elements of $Perm_1$ and $Perm_2$.

$(sp2)$ Weak structural property (\textit{cardinality of ordered focal elements}):
A distance measure $d$ between two permutation mass functions $Perm_1$ and $Perm_2$ is considered \textit{weakly structural} if its formulation incorporates the cardinality of ordered focal elements of $Perm_1$ and $Perm_2$.

$(sp3)$ Order structural property (\textit{order of ordered focal elements}):
A distance measure $d$ between two permutation mass functions $Perm_1$ and $Perm_2$ is considered \textit{order structural} if its formulation incorporates the order of ordered focal elements of $Perm_1$ and $Perm_2$.

 $(sp4)$ Order structural dissimilarity (\textit{interaction between sets of ordered focal elements}):
A distance measure $d$ between two permutation mass functions $Perm_1$ and $Perm_2$ is considered to satisfy the \textit{order structurally dissimilarity} if its formulation incorporates the interactions between the sets $\mathcal{OF}_1$ and $\mathcal{OF}_2$ of ordered focal elements of $Perm_1$ and $Perm_2$. 

Table \ref{tab5} summarizes the metric and structural properties of the RPS distance measure method using different weighting matrices. We do not claim that any particular distance measure is universally superior; instead, we aim to provide applicability guidance. For instance, if the goal is to measure the distance between layer-2 belief structure information distributions, the measure $d_{{CD}^{[Orn]}}$ may be an appropriate choice. If you are additionally pursuing the full metric, you can choose the corrected metric  $d_{{CD}^{[Orn, \lambda_{min}]}}$.

 \begin{table}[!htbp]\small
    	\begin{center}
    		\setlength{\tabcolsep}{0.5cm}
    		\caption{Metric and structural properties of RPS distance measures using different weighting matrices}
    		\begin{tabular}{p{3em}p{1em}p{1em}p{1em}p{1em}p{1em}p{1em}p{1em}p{1em}p{1em}}
    			\toprule
                     $Distance$&  \multicolumn{5}{c}{$Metric$} &  \multicolumn{4}{c}{$Structural$}  \\
                    \cmidrule(lr){2-6}\cmidrule(lr){7-10} 
             			  &$(d1)$& $(d2)$ & $(d3)'$ & $(d3)''$ & $(d4)$&$(sp1)$&$(sp2)$&$(sp3)$&$(sp4)$  \\
    			\midrule
    			\specialrule{0em}{1pt}{1pt}
    			 $d_{D}$\cite{jousselme2001new}  &  & $\surd$ & $\surd$& & &$\surd$ & & &   \\
        		\specialrule{0em}{1pt}{1pt}
    			 $d_{RD}$\cite{chen2023distance}  &  & $\surd$ & $\surd$& & & $\surd$&  &$\surd$ &   \\
        		\specialrule{0em}{1pt}{1pt}
    			 $d_{{CD}^{[Orn]}}$   &  & $\surd$ & $\surd$& & &  $\surd$ &  &$\surd$  &  \\
        		\specialrule{0em}{1pt}{1pt}  
    			 $d_{{CD}^{[Orn, \lambda_{min}]}}$ & $\surd$  & $\surd$ & $\surd$ & $\surd$ & $\surd$ & $\surd$ &  &$\surd$  &  \\
                    \specialrule{0em}{1pt}{1pt}
    			\bottomrule
    			\label{tab5}
    		\end{tabular}
    	\end{center}
    \end{table}

\section{Numerical Examples and Discussion}\label{discussion}
In this section, we present numerical examples to illustrate the behavior and characteristics of the proposed RPS distance measure. The advantages of the proposed method are highlighted through a comparative analysis with the existing RPS distance measure proposed by Chen et al. \cite{chen2023distance}. Finally, the effect of the weight setting on the distance of RPSs is discussed.

\subsection{Numerical examples}

$\mathbf{Example \ 5.1.}$  Suppose there are two RPS defined on a fixed set $\mathcal{T}=\{\tau_1, \tau_2, \tau_3, \tau_4, \tau_5\}$, and their PMFs are as follows
\begin{equation}
\begin{aligned}
&PMF_1(\mathcal{T}): Perm_1(F_{7}^1)=1 \ \ \  \Rightarrow  \ \   Perm_1((\tau_1\tau_2\tau_3))=1;\\
&PMF_2(\mathcal{T}): Perm_2(F_{24}^1)=1 \ \ \   \Rightarrow  \ \   Perm_2((\tau_4\tau_5))=1. \notag
\end{aligned}
\end{equation}

Both the proposed method and the existing RPS distance measure\cite{chen2023distance} obtain an intuitive result of $d_{RPS}(Perm_1, Perm_2)=1$.

$\mathbf{Example \ 5.2.}$ Consider two RPS defined on a 3-element FoD $\mathcal{T}=\{\tau_1, \tau_2, \tau_3\}$ as follows
\begin{equation}
\begin{aligned}
&PMF_1(\mathcal{T}): Perm_1(F_{1}^1)=\frac{1}{4}, \ \ \ \ Perm_1(F_{2}^1)=\frac{1}{2}, \ \ \ \ Perm_1(F_{4}^1)=\frac{1}{4};   \\
&\ \ \ \ \ \ \Rightarrow  \ \ \ \ \ Perm_1((\tau_1))=\frac{1}{4},\ \ \ Perm_1((\tau_2))=\frac{1}{2},\ \ \ Perm_1((\tau_3))=\frac{1}{4}.  \\
&PMF_2(\mathcal{T}): Perm_2(F_{1}^1)=Perm_2(F_{2}^1)=Perm_2(F_{4}^1)=\frac{1}{3};              \\
&\ \ \ \ \ \ \Rightarrow  \ \ \ \ \ Perm_2((\tau_1))=Perm_2((\tau_2))=Perm_2((\tau_3))=\frac{1}{3}.   \notag
\end{aligned}
\end{equation}

When beliefs are only assigned to the singleton, the PMF can be classified as a Bayesian mass function. The corresponding mass functions are $m_1(\{\tau_1\})=\frac{1}{4}, m_1(\{\tau_2\})=\frac{1}{2}, m_1(\{\tau_3\})=\frac{1}{4}$ and $m_2(\{\tau_1\})=m_2(\{\tau_2\})=m_2(\{\tau_3\})=\frac{1}{3}$ respectively. In this case, the proposed RPS distance measure degenerates into the normalized Euclidean distance $d_{E}$ between the probability distributions. Assuming there are discrete probability distributions $P=[\frac{1}{4}, \frac{1}{2}, \frac{1}{4}]$ and $Q=[\frac{1}{3}, \frac{1}{3}, \frac{1}{3}]$, we have
$$d_{RPS}(Perm_1, Perm_2)=d_{BPA}(m_1, m_2)=d_{E}(P, Q)=0.1443.$$
The same results can be obtained using the distance measure method proposed by Chen et al.\cite{chen2023distance}.

$\mathbf{Example \ 5.3.}$ Following Example 3.1, given three RPSs from distinct sources on the frame $\mathcal{T}=\{\tau_1, \tau_2, \tau_3\}$ be as follows:
\begin{equation}
\begin{aligned}
&PMF_1(\mathcal{T}): Perm_1(F_3^1)=1\ \ \Rightarrow\ Perm_1((\tau_1\tau_2))=1, \\
&PMF_2(\mathcal{T}): Perm_2(F_7^1)=1\ \ \Rightarrow\ Perm_2((\tau_1\tau_2\tau_3))=1, \\
&PMF_3(\mathcal{T}): Perm_3(F_7^5)=1\ \ \Rightarrow\ Perm_3((\tau_3\tau_1\tau_2))=1. \notag
\end{aligned}
\end{equation}

The distances among $RPS_1$, $RPS_2$, and $RPS_3$ are calculated using both the existing RPS distance measure method\cite{chen2023distance} and the proposed method, as presented in Table \ref{tab6}. In Section \ref{Different interpretations}, we focused solely on the relationship between $d_{RPS}(Perm_1, Perm_2)$ and $d_{RPS}(Perm_1, Perm_3)$ under different interpretations of information distribution, aiming to illustrate that distinct interpretations lead to different behaviors of the distances. From the perspective of RFS, these two distances are equal, while under the TBM interpretation, the former is smaller than the latter. Here, we will further elaborate on the relationships among $d_{RPS}(Perm_1, Perm_2)$, $d_{RPS}(Perm_1, Perm_3)$ and $d_{RPS}(Perm_2, Perm_3)$ within layer-2 belief structure. 

\begin{table}[!htbp]
    	\begin{center}
    		\setlength{\tabcolsep}{0.5cm}
    		\caption{The distances among $RPS_1$, $RPS_2$, and $RPS_3$ in Example 5.3}
    		\begin{tabular}{ccc}
    			\toprule
    			  & Chen et al.\cite{chen2023distance} &Proposed method   \\
    			\midrule
    			\specialrule{0em}{1pt}{1pt}
    			$d_{RPS}(Perm_1,Perm_2)$ &0.5774 &0.3333   \\
        		\specialrule{0em}{1pt}{1pt}
    			$d_{RPS}(Perm_1,Perm_3)$ &0.8110  &0.8165  \\
        		\specialrule{0em}{1pt}{1pt}
    			$d_{RPS}(Perm_2,Perm_3)$  &0.8581 &0.7454 \\
    			\bottomrule
    			\label{tab6}
    		\end{tabular}
    	\end{center}
    \end{table}
    
From Table \ref{tab6}, it can be observed that the distances between $RPS_1$, $RPS_2$, and $RPS_3$, as computed by the proposed method, follow the relationship $d_{RPS}(Perm_1, Perm_2)< d_{RPS}(Perm_2, Perm_3)< d_{RPS}(Perm_1, Perm_3)$. Under the TBM interpretation, there exists an inconsistency in qualitative propensity between $Perm_1$ and $Perm_3$, as well as between $Perm_2$ and $Perm_3$. Since $PMF_1$ and $PMF_2$ share a similar belief transfer propensity, $\tau_1\succ\tau_2$, the degree of inconsistency in qualitative propensity caused by $\tau_1$ and $\tau_2$ remains the same when comparing the distances of $Perm_1$ and $Perm_2$ relative to $Perm_3$, respectively. Further analysis shows that the inconsistency caused by the target element $\tau_3$ between $Perm_2$ and $Perm_3$ arises from the different preferences of $PMF_2$ and $PMF_3$ for $\tau_3$, whereas that between $Perm_1$ and $Perm_3$ stems from the absence of $\tau_3$ in $PMF_1$. Compared to $d_{RPS}(Perm_2, Perm_3)$, which is only attributed to the inconsistent qualitative propensity, there is additionally an inconsistency in the number and type of target elements considered by the agent in $d_{RPS}(Perm_1, Perm_3)$. Consequently, $d_{RPS}(Perm_2, Perm_3)< d_{RPS}(Perm_1, Perm_3)$. From the TBM perspective, inconsistency in the preferences assigned to an included element is expected to have a smaller impact than the difference between including and entirely excluding the element.

However, the result obtained by Chen et al.' distance measure method\cite{chen2023distance} is $d_{RPS}(Perm_1, Perm_2)< d_{RPS}(Perm_1, Perm_3)< d_{RPS}(Perm_2, Perm_3)$, which does not conform to either the RFS or TBM interpretation. The reason for this unreasonable result is that the ordered degree of permutation events, as defined in the distance measure of Chen et al.\cite{chen2023distance}, only considers the position difference between the common elements of two permutations. The permutation events $F_3^1$ in $PMF_1$ and $F_7^5$ in $PMF_3$ share two common elements, while $F_7^1$ in $PMF_2$ and $F_7^5$ in $PMF_3$ share three common elements. Since the positions of the element $\tau_1,\tau_2$ in $F_3^1$ and $F_7^1$ are the same, the additional common element $\tau_3$ increases the distance between $Perm_2$ and $Perm_3$, resulting in $ d_{RPS}(Perm_1, Perm_3)< d_{RPS}(Perm_2, Perm_3)$. However, they indeed neither provided a reasonable explanation for such results nor the applicable scenarios for their distance measure in literature  \cite{chen2023distance}.

$\mathbf{Example \ 5.4.}$ Consider two RPSs defined on an 8-element FoD, whose PMFs are
\begin{equation}
\begin{aligned}
&PMF_1(\mathcal{T}): Perm_1(F_{32}^1)=0.2, \ \ \ \ Perm_1(F_{192}^1)=0.3, \ \ \ \ Perm_1(F_{31}^1)=0.5;   \\
&\ \ \ \ \ \ \Rightarrow  \ \ \ \ \  Perm_1((\tau_6))=0.2,\ Perm_1((\tau_7\tau_8))=0.3,\ Perm_1((\tau_1\tau_2\tau_3\tau_4\tau_5))=0.5.   \\
&PMF_2(\mathcal{T}): Perm_2((X))=1.                   \notag
\end{aligned}
\end{equation}
When the order of ordered focal sets is not taken into account, PMFs degenerate into mass functions in DST as follows.
\begin{equation}
\begin{aligned}
&  m_1(\mathcal{T}):  \  m_1(F_{32})=0.2, \ \ \ \ m_1(F_{192})=0.3, \ \ \ \ m_1(F_{31})=0.5;   \\
&  \ \  \Rightarrow  \ \ \  \ \  m_1(\{\tau_6\})=0.2,\ \ m_1(\{\tau_7, \tau_8\})=0.3,\ \ m_1(\{\tau_1, \tau_2, \tau_3, \tau_4, \tau_5\})=0.5   \\
&  m_2(\mathcal{T}): \   m_2(\{X\})=1;                   \notag
\end{aligned}
\end{equation}

\begin{figure}[htbp!]
    \centering
    \includegraphics[width=0.6\textwidth]{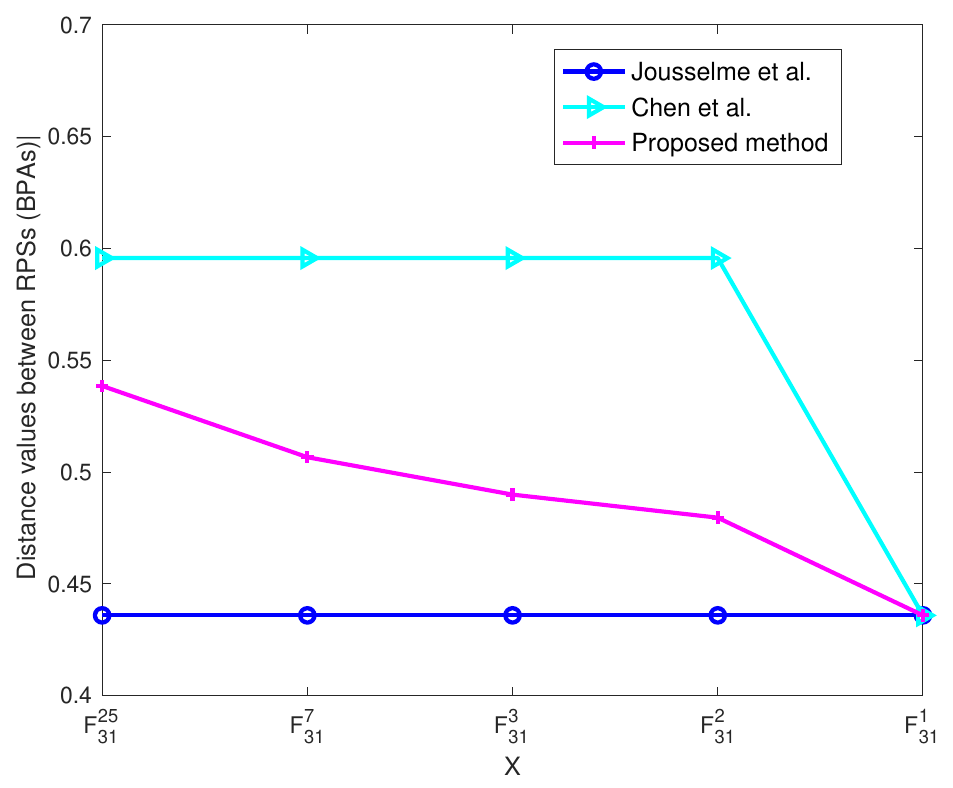}
    \caption{The distance between RPSs(BPAs) when $X$ takes different values in Example 5.4}
    \label{Example4.4}
\end{figure}

Table \ref{tab7} and Figure \ref{Example4.4} present the distances between $RPS_1$ and $RPS_2$ calculated using the proposed method and the existing RPS distance measure method\cite{chen2023distance} for different values of $X$. Additionally, the distance between the corresponding mass functions $m_1$ and $m_2$ is provided based on the Jousselme distance\cite{jousselme2001new}.

 \begin{table}[!htbp]
    	\begin{center}
    		\setlength{\tabcolsep}{0.5cm}
    		\caption{The distance between RPSs(BPAs) when $X$ takes different values in Example 5.4}
    		\begin{tabular}{cccc}
    			\toprule
                     $X$&  $d_{BPA}$& \multicolumn{2}{c}{$d_{RPS}$}  \\
                    \cmidrule(lr){3-4}
    			  &Jousselme et al.\cite{jousselme2001new}& Chen et al.\cite{chen2023distance} &Proposed method   \\
    			\midrule
    			\specialrule{0em}{1pt}{1pt}
    			$F_{31}^{25}: \tau_2\tau_1\tau_3\tau_4\tau_5$ &0.4359 &0.5957  &0.5385 \\
        		\specialrule{0em}{1pt}{1pt}
    			$F_{31}^{7}: \tau_1\tau_3\tau_2\tau_4\tau_5$ &0.4359  &0.5957 &0.5066  \\
        		\specialrule{0em}{1pt}{1pt}
    			$F_{31}^{3}: \tau_1\tau_2\tau_4\tau_3\tau_5$  &0.4359 &0.5957 &0.4899 \\
        		\specialrule{0em}{1pt}{1pt}
    			$F_{31}^{2}: \tau_1\tau_2\tau_3\tau_5\tau_4$  &0.4359  &0.5957 &0.4796  \\
                    \specialrule{0em}{1pt}{1pt}
    			$F_{31}^{1}: \tau_1\tau_2\tau_3\tau_4\tau_5$ & 0.4359 &0.4359 &0.4359 \\
    			\bottomrule
    			\label{tab7}
    		\end{tabular}
    	\end{center}
    \end{table}

It can be seen that $X$ follows five distinct orderings of the focal elements $\tau_1, \tau_2, \tau_3, \tau_4$ and $\tau_5$, each corresponding to a different permutation event. These five orders are based on $(\tau_1\tau_2\tau_3\tau_4\tau_5)$ and are generated sequentially by perturbing two elements from the front to the back. Observe the distance values obtained by the proposed method. When the top two elements $\tau_1$ and $\tau_2$ are swapped, the distance between $RPS_1$ and $RPS_2$ is the largest at 0.5385. As the order inconsistency occurs in the elements ranked later and later, the distance gradually decreases, reaching 0.4796 when the last two elements are swapped. When the order of $X$ aligns with $F_{31}^1\in \mathcal{OF}_1$, the distance reaches the minimum value of 0.4359.

For the corresponding mass function, $X$ degenerates to the unordered focal set $F_{31}:\{\tau_1, \tau_2, \tau_3, \tau_4, \tau_5 \}$. At this time, the change in the order of elements in $X$ has no effect on the distance between $m_1$ and $m_2$, so $d_{BPA}$ remains unchanged at 0.4359, which is equal to the distance between RPSs when the order of $X$ is consistent with $F_{31}^1\in \mathcal{OF}_1$.

However, the results of the existing RPS distance measure proposed by Chen et al.\cite{chen2023distance} can only distinguish the disturbed and undisturbed order cases. The distance obtained when the order is not disturbed is 0.4359, the same as the proposed method and the Jousselme distance\cite{jousselme2001new}; when two elements are disturbed, whether the top or the last two elements, the distance obtained is 0.5957 without any distinction. Compared to the existing RPS distance measure\cite{chen2023distance}, the proposed method can more effectively reflect the impact of the element order in permutation events on the distance between PMFs. It exhibits a natural top-weightiness property, meaning that top-ranked elements occupy a more critical position. In other words, inconsistent order of higher-ranked elements will lead to a larger distance between RPSs, which aligns with the layer-2 belief structure interpretation of RPS.

$\mathbf{Example \ 5.5}$ Assume there are three RPSs defined on a frame with 10 elements, and their PMFs are as follows
\begin{equation}
\begin{aligned}
&PMF_1(\mathcal{T}): Perm_1(F_{8}^1)=0.05, \ \ \ \ Perm_1(F_{5}^1)=0.05, \ \ \ \ Perm_1((X))=0.8, \\
&\ \ \ \ \ \ \ \ \ \ \ \ \ \ \ \  Perm_1(F_{31}^1)=0.1;   \\
&\ \ \ \ \ \ \Rightarrow \ \ \ \ \  Perm_1((\tau_4))=0.05,\ \ Perm_1((\tau_2\tau_3))=0.05,\ \ Perm_1((X))=0.8,\\
& \ \ \ \ \ \ \  \ \ \ \ \ \ \ \ \  Perm_1((\tau_1\tau_2\tau_3\tau_4\tau_5))=0.1  \notag\\
&PMF_2(\mathcal{T}): Perm_2(F_7^1)=1.              \\
&\ \ \ \ \ \ \Rightarrow \ \ \ \ \ Perm_2((\tau_1\tau_2\tau_3))=1.  \\
&PMF_3(\mathcal{T}): Perm_3(F_7^6)=1.              \\
&\ \ \ \ \ \ \Rightarrow \ \ \ \ \ Perm_3((\tau_3\tau_2\tau_1))=1.  \notag
\end{aligned}
\end{equation}
If the order of ordered focal sets is disregarded, PMFs degenerate into mass functions in DST as follows.
\begin{equation}
\begin{aligned}
&m_1(\mathcal{T}): m_1(F_{8})=0.05, \  m_1(F_{5})=0.05, \ m_1(\{X\})=0.8,\ m_1(F_{31})=0.1;\\
& \ \ \ \Rightarrow \ \ \  m_1(\{\tau_4\})=0.05,\ \ m_1(\{\tau_2, \tau_3\})=0.05,\ \ m_1(\{X\})=0.8,\\
& \ \ \ \ \ \ \  \ \ \ \ m_1(\{\tau_1, \tau_2, \tau_3, \tau_4, \tau_5\})=0.1  \notag\\
&m_2(\mathcal{T}): m_2(F_7)=1.              \\
&\ \ \ \Rightarrow \ \ \  m_2(\{\tau_1, \tau_2, \tau_3\})=1.  \\
&m_3(\mathcal{T}): m_2(F_7)=1.              \\
&\ \ \ \Rightarrow \ \ \  m_3(\{\tau_1, \tau_2, \tau_3\})=1.  \notag
\end{aligned}
\end{equation}

\begin{figure}[htbp!]
    \centering
    \includegraphics[width=0.6\textwidth]{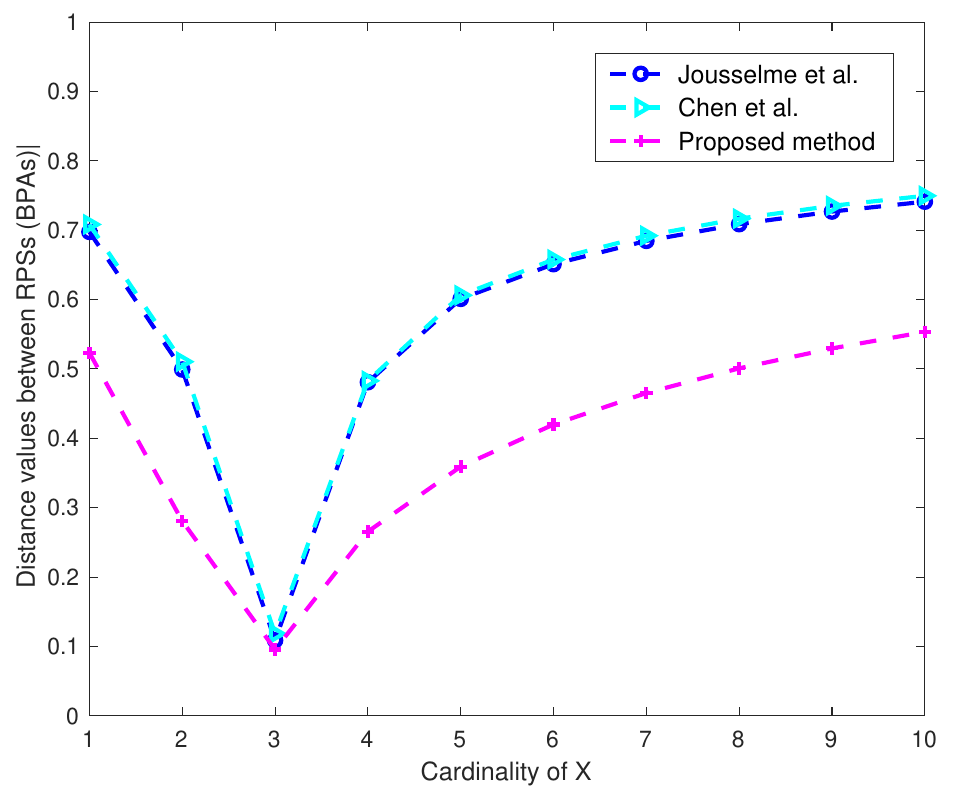}
    \caption{The distance between $RPS_1$ and $RPS_2$($BPA_1$ and $BPA_2$) when $X$ changes from $\tau_1$ to $\tau_1\tau_2\tau_3...\tau_{10}$ in Example 5.5}
    \label{Example4.5a}
\end{figure}

Let us first focus on the distance between $RPS_1$ and $RPS_2$. Figure \ref{Example4.5a} and the upper part of Table \ref{tab8} show the distance between $RPS_1$ and $RPS_2$ obtained by the proposed method and Chen et al.' method\cite{chen2023distance}, along with the Jousselme distance\cite{jousselme2001new} between their corresponding mass functions, when $X$ changes from $\tau_1$ to $\tau_1\tau_2\tau_3...\tau_{10}$.

\begin{table}[!htbp]\tiny
    	\begin{center}
        	\setlength{\tabcolsep}{0.5cm}
    		\caption{The distance between RPSs (BBAs) when $X$ changes in Example 5.5}
    		\begin{tabular}{lccc}
    			\toprule
                            $X$&  $d_{BPA}(m_1, m_2)$& \multicolumn{2}{c}{$d_{RPS}(Perm_1, Perm_2)$}  \\
                   \cmidrule(lr){3-4}
                 &Jousselme et al.\cite{jousselme2001new}& Chen et al.\cite{chen2023distance} &Proposed method   \\
    			\midrule
    			\specialrule{0em}{1pt}{1pt}
    			$F_1^1: \tau_1$ & 0.6975 &0.7082  &0.5233 \\
        		\specialrule{0em}{1pt}{1pt}
    			$F_3^1:\tau_1\tau_2$ &0.4992  &0.5104 &0.2807 \\
        		\specialrule{0em}{1pt}{1pt}
    			$F_7^1:\tau_1\tau_2\tau_3$  & 0.1088  &0.1182 &0.0955   \\
        		\specialrule{0em}{1pt}{1pt}
    			$F_{15}^1:\tau_1\tau_2\tau_3\tau_4$  & 0.4808 &0.4828 &0.2655 \\
        		\specialrule{0em}{1pt}{1pt}
    			$F_{31}^1:\tau_1\tau_2\tau_3\tau_4\tau_5$  & 0.6010 &0.6061 &0.3591\\
         		\specialrule{0em}{1pt}{1pt}
    			$F_{63}^1:\tau_1\tau_2\tau_3\tau_4\tau_5\tau_6$  & 0.6510 &0.6577  &0.4199  \\
         		\specialrule{0em}{1pt}{1pt}
    			$F_{127}^1:\tau_1\tau_2\tau_3\tau_4\tau_5\tau_6\tau_7$  & 0.6845  &0.6920 &0.4635  \\
         		\specialrule{0em}{1pt}{1pt}
    			$F_{255}^1:\tau_1\tau_2\tau_3\tau_4\tau_5\tau_6\tau_7\tau_8$  & 0.7086  &0.7166 &0.5008   \\
         		\specialrule{0em}{1pt}{1pt}
    			$F_{511}^1:\tau_1\tau_2\tau_3\tau_4\tau_5\tau_6\tau_7\tau_8\tau_9$  & 0.7268  &0.7351 &0.5296  \\
         		\specialrule{0em}{1pt}{1pt}
    			$F_{1023}^1:\tau_1\tau_2\tau_3\tau_4\tau_5\tau_6\tau_7\tau_8\tau_9\tau_{10}$  &0.7411 & 0.7495  &0.5534   \\
    			\toprule
                                   $X$&  $d_{BPA}(m_1, m_3)$& \multicolumn{2}{c}{$d_{RPS}(Perm_1, Perm_3)$}  \\
                   \cmidrule(lr){3-4}
                 &Jousselme et al.\cite{jousselme2001new}& Chen et al.\cite{chen2023distance} &Proposed method   \\
    			\midrule
    			\specialrule{0em}{1pt}{1pt}
    			$F_1^1: \tau_1$ & 0.6975 &0.8151  &0.8343 \\
        		\specialrule{0em}{1pt}{1pt}
    			$F_3^2:\tau_2\tau_1$ &0.4992  &0.7391 &0.7413 \\
        		\specialrule{0em}{1pt}{1pt}
    			$F_7^6:\tau_3\tau_2\tau_1$  & 0.1088  &0.1434 &0.1165   \\
        		\specialrule{0em}{1pt}{1pt}
    			$F_{15}^{24}:\tau_4\tau_3\tau_2\tau_1$  & 0.4808 &0.7328&0.7102 \\
        		\specialrule{0em}{1pt}{1pt}
    			$F_{31}^{120}:\tau_5\tau_4\tau_3\tau_2\tau_1$  & 0.6010 &0.8135 &0.7776\\
         		\specialrule{0em}{1pt}{1pt}
    			$F_{63}^{720}:\tau_6\tau_5\tau_4\tau_3\tau_2\tau_1$  & 0.6510 &0.8446  &0.8075  \\
         		\specialrule{0em}{1pt}{1pt}
    			$F_{127}^{5040}:\tau_7\tau_6\tau_5\tau_4\tau_3\tau_2\tau_1$  & 0.6845  &0.8595 &0.8249  \\
         		\specialrule{0em}{1pt}{1pt}
    			$F_{255}^{40320}:\tau_8\tau_7\tau_6\tau_5\tau_4\tau_3\tau_2\tau_1$  & 0.7086  &0.8667 &0.8351   \\
         		\specialrule{0em}{1pt}{1pt}
    			$F_{511}^{362880}:\tau_9\tau_8\tau_7\tau_6\tau_5\tau_4\tau_3\tau_2\tau_1$  & 0.7268  &0.8712 &0.8420  \\
         		\specialrule{0em}{1pt}{1pt}
    			$F_{1023}^{3628800}:\tau_{10}\tau_9\tau_8\tau_7\tau_6\tau_5\tau_4\tau_3\tau_2\tau_1$  &0.7411 & 0.8737  &0.8464  \\
    			\bottomrule
    			\label{tab8}
    		\end{tabular}
    	\end{center}
    \end{table}

We can observe from Figure \ref{Example4.5a} and Table \ref{tab8} that the results obtained by the proposed method show an intuitive trend. When the cardinality of $X$ gets closer to 3, the distance between $RPS_1$ and $RPS_2$ decreases. On the contrary, as $|X|$ deviates further from 3, the distance increases. Moreover, it is worth noting that when the inconsistency between $X$ and $F_7^1$ arises from one unconsidered element, the distance between $RPS_1$ and $RPS_2$ is greater for $|X|=2$ than for $|X|=4$. Similarly, when two unconsidered elements cause the inconsistency between $X$ and $F_7^1$, the distance when $|X|=1$ is more significant than when $|X|=5$ or even $|X|=8$. This phenomenon precisely reflects the top-weightiness nature of the proposed distance measure, wherein inconsistencies among top-ranked elements lead to larger distances. The same conclusion can be reached by the distance measure method proposed by Chen et al.\cite{chen2023distance}.

\begin{figure}[htbp!]
    \centering
    \includegraphics[width=0.6\textwidth]{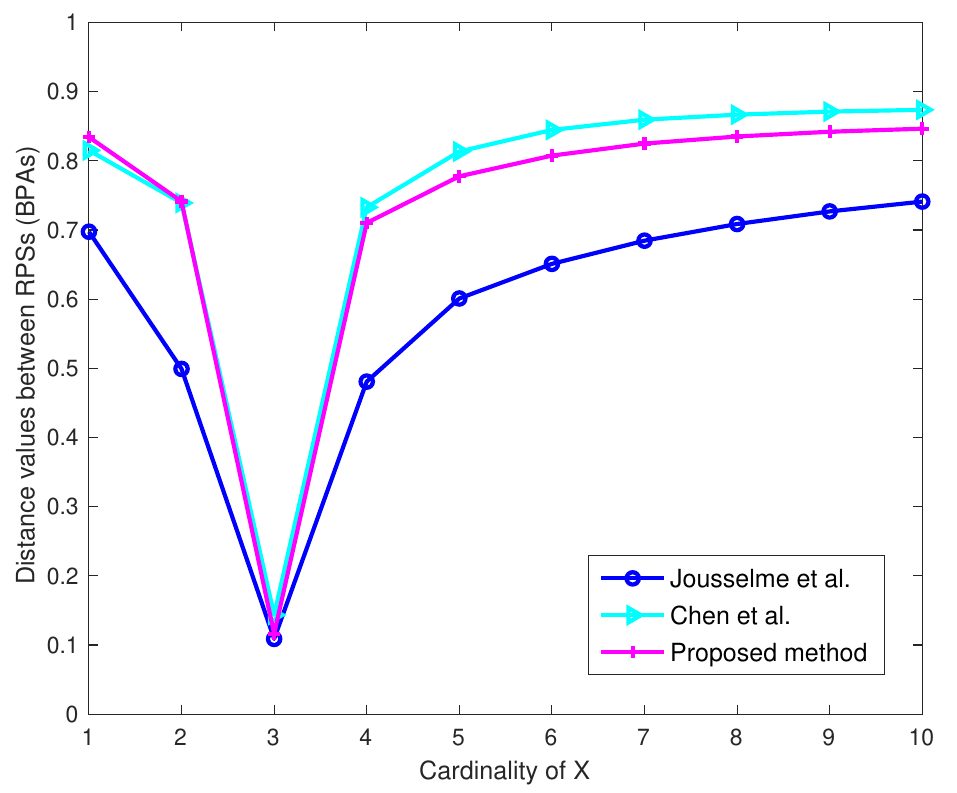}
    \caption{The distance between $RPS_1$ and $RPS_3$($BPA_1$ and $BPA_3$) when $X$ takes values from $\tau_1$ to $\tau_{10}\tau_9\tau_8...\tau_1$ in reverse order in Example 5.5}
    \label{Example4.5b}
\end{figure}

\begin{figure}[htbp!]
    \centering
    \includegraphics[width=0.6\textwidth]{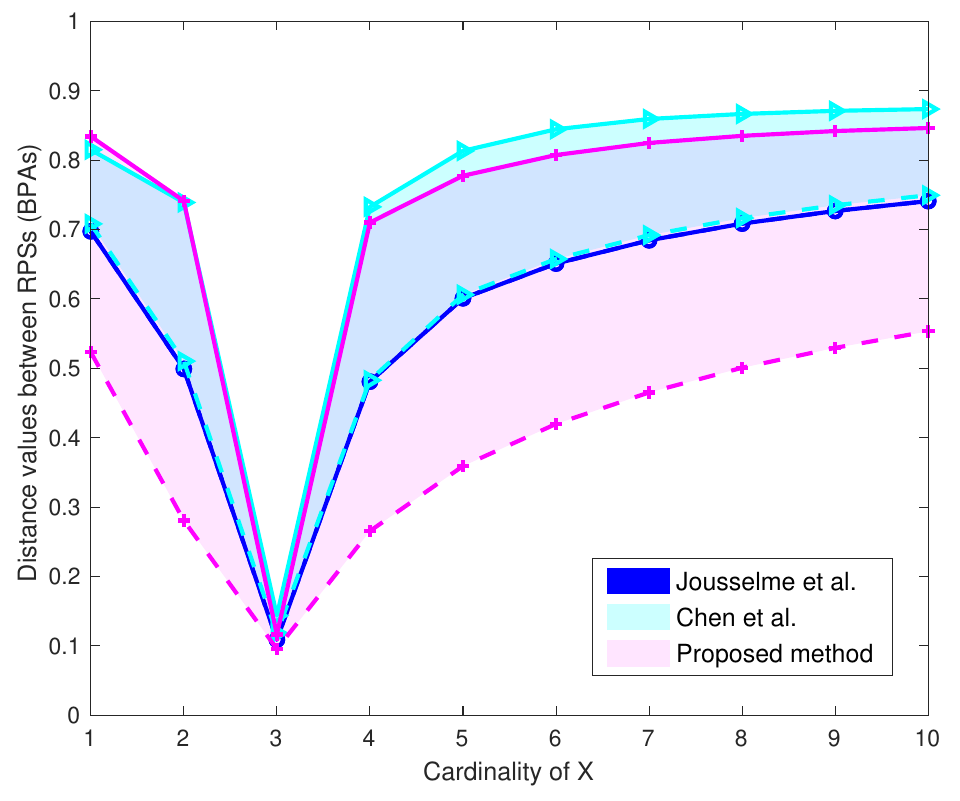}
    \caption{Distance between RPSs(BBAs) when $X$ takes different values in Example 5.5}
    \label{Shadow}
\end{figure}

Introducing $RPS_3$, the distance between $RPS_1$ and $RPS_3$ obtained by the proposed method and Chen et al.' method\cite{chen2023distance}, as well as the Jousselme distance\cite{jousselme2001new} between their corresponding mass functions, when $X$ takes values from $\tau_1$ to $\tau_{10}\tau_9\tau_8...\tau_1$ in reverse order, are shown in Figure \ref{Example4.5b} and the lower part of Table \ref{tab8}. Different from the above case, the change in the cardinality of $X$ here increases the inconsistency between the top-ranked elements. Consequently, the distance between $RPS_1$ and $RPS_3$ is generally greater than that between $RPS_1$ and $RPS_2$. To enhance clarity, Figures \ref{Example4.5a} and \ref{Example4.5b} are combined into Figure \ref{Shadow} for better visualization. It is easy to find that the proposed method presents a larger shadow area in Figure \ref{Shadow}, which indicates that compared with Chen et al.' method\cite{chen2023distance}, the proposed method is more sensitive to the change of the order of focal elements in the ordered focal set when measuring the distance between two RPSs.

For the corresponding mass function, the order of the ordered focus is no longer considered. At this time, $m_2=m_3$, so the results obtained by the Jousselme distance in Figure \ref{Shadow} show an overlapping line.

$\mathbf{Example \ 5.6.}$ Consider two RPSs defined on a fixed set $\mathcal{T}=\{\tau_1, \tau_2, \tau_3, \tau_4, \tau_5, \tau_6, \tau_7\}$, and their PMFs are as follows
\begin{equation}
\begin{aligned}
&PMF_1(\mathcal{T}): Perm_1(F_{31}^{31})=0.4, \ \ \ \ Perm_1(F_{6}^1)=0.6;  \\
&\ \ \ \ \ \ \Rightarrow \ \ \ \ \  Perm_1((\tau_2\tau_3\tau_1\tau_4\tau_5))=0.4,\ \ Perm_1((\tau_2\tau_3))=0.6;\\
&PMF_2(\mathcal{T}): Perm_2(F_{127}^{841})=1.              \\
&\ \ \ \ \ \ \Rightarrow \ \ \ \ \ Perm_2((\tau_2\tau_3\tau_1\tau_4\tau_5\tau_6\tau_7))=1.  \notag
\end{aligned}
\end{equation}
When the order of ordered focal sets is not taken into account, PMFs degenerate into mass functions in DST as follows.
\begin{equation}
\begin{aligned}
&m_1(\mathcal{T}): m_1(F_{31})=0.4, \  m_1(F_{6})=0.6;\\
&\ \ \ \Rightarrow \ \ \  m_1(\{\tau_1, \tau_2, \tau_3, \tau_4, \tau_5\})=0.4\ \ m_1(\{\tau_2, \tau_3\})=0.6. \\
&m_2(\mathcal{T}): m_2(F_{127})=1.              \\
&\ \ \ \Rightarrow \ \ \  m_2(\{\tau_1, \tau_2, \tau_3, \tau_4, \tau_5, \tau_6, \tau_7\})=1.  \notag
\end{aligned}
\end{equation}

\begin{figure}[htbp!]
    \centering
    \includegraphics[width=0.6\textwidth]{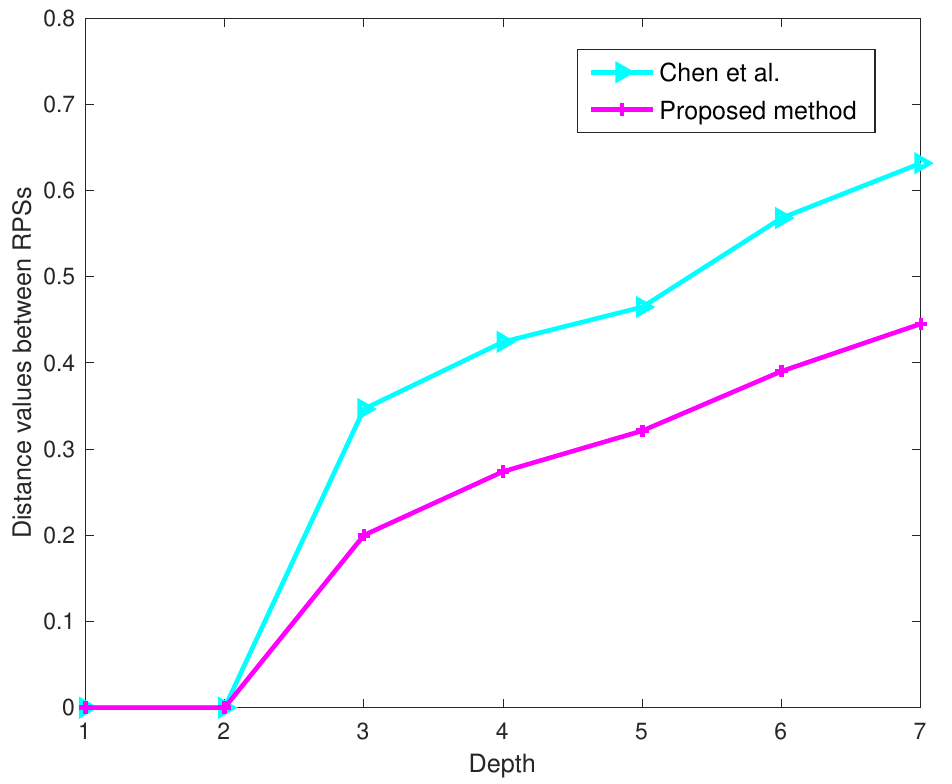}
    \caption{The distance between RPSs at different depths in Example 5.6}
    \label{Example4.6}
\end{figure}

The distances between $RPS_1$ and $RPS_2$ at different depths, calculated using both the proposed method and the existing RPS distance measure\cite{chen2023distance}, are given in Table \ref{tab9} and Figure \ref{Example4.6}. Since the focal set is unordered in the mass function, indicating that its elements are of equal status and are treated equally, it is not very sensible to consider the distances between mass functions at different depths. $d_{BPA}$ is not applicable in this case. Although the distance measure proposed by Chen et al.\cite{chen2023distance} does not explicitly include a parameter $t$ for adjusting the depth, we still provide the distances between the PMFs of $RPS_1$ and $RPS_2$ at different depths as a reference for comparison. For example, $RPS_1$ and $RPS_2$ at depth 3 correspond to the new PMFs $Perm_1'((\tau_2\tau_3\tau_1))=0.4, Perm_1'((\tau_2\tau_3))=0.6$ and $Perm_2'((\tau_2\tau_3\tau_1))=1$ under the same FoD, which is essentially different from the use of parameter $t$ in the proposed method to control the different truncation depths of RPSs.

\begin{table}[!htbp]
    	\begin{center}
        	\setlength{\tabcolsep}{0.5cm}
    		\caption{The distance between RPSs at different depths in Example 5.6}
    		\begin{tabular}{lccc}
    			\toprule
    			$Depth$ & $d_{BPA}$ &  \multicolumn{2}{c}{$d_{RPS}$} \\
                    \cmidrule(lr){3-4}
                            &Jousselme et al.\cite{jousselme2001new}& Chen et al.\cite{chen2023distance} &Proposed method   \\
    			\midrule
    			\specialrule{0em}{1pt}{1pt}
    			$t=1$ & - &0.0000  &0.0000 \\
        		\specialrule{0em}{1pt}{1pt}
    			$t=2$ &-  &0.0000 &0.0000 \\
        		\specialrule{0em}{1pt}{1pt}
    			$t=3$  & -  &0.3464 &0.2000  \\
        		\specialrule{0em}{1pt}{1pt}
    			$t=4$ & - &0.4243  &0.2739 \\
        		\specialrule{0em}{1pt}{1pt}
    			$t=5$ & -  &0.4648 &0.3212 \\
         		\specialrule{0em}{1pt}{1pt}
    			$t=6$ & - &0.5680  &0.3903\\
         		\specialrule{0em}{1pt}{1pt}
    			$t=7$ & - &0.6316 &0.4453  \\
    			\bottomrule
    			\label{tab9}
    		\end{tabular}
    	\end{center}
    \end{table}

By observing Table \ref{tab9} and Figure \ref{Example4.6}, it is evident that the proposed method can effectively measure the distance between arbitrarily truncated RPSs.  $PMF_1$ indicates that the belief of 0.4 exhibits a transfer propensity among the target elements in the order of $\tau_2\succ\tau_3\succ\tau_1\succ\tau_4\succ\tau_5$, while a belief of 0.6 follows the order $\tau_2\succ\tau_3$. Similarly, $PMF_2$ indicates that the target elements exhibit a transfer propensity of $\tau_2\succ\tau_3\succ\tau_1\succ\tau_4\succ\tau_5\succ\tau_6\succ\tau_7$ with a belief of 1. When considering only the top two target elements, $RPS_1$ and $RPS_2$ show the same qualitative propensity, resulting in $d_{RPS}(Perm_1, Perm_2)=0$. As attention shifts to the top three or top five target elements, $F_{127}^{841}\in\mathcal{OF}_2$ and $F_{32}^{31}\in\mathcal{OF}_1$ still show the same belief transfer propensity, whereas the inconsistency between $F_{127}^{841}\in\mathcal{OF}_2$ and $F_{6}^{1}\in\mathcal{OF}_1$ gradually emerges, leading to an increasing distance. When the depth of attention reaches 6 or 7, the inconsistency between $F_{127}^{841}\in\mathcal{OF}_2$ and $F_{32}^{31}\in\mathcal{OF}_1$ begins to appear, and the distance further increases. In particular, the trend presented by the proposed method is consistent with the distance between RPSs corresponding to different depths using Chen et al.' method\cite{chen2023distance}. Generally speaking, for two RPSs, as the depth increases, the possibility of inconsistency between the ordered focal sets $\mathcal{OF}_1$ and $\mathcal{OF}_2$ will also increase. The proposed distance measure method offers decision-makers flexibility, allowing them to concentrate solely on the aspects of interest.

\subsection{Discussion on different weights}
Above, we discussed the case of $Orn = 0.5$, where each depth is assigned the same weight, namely $\mathbf{w}^{[0.5]}=[\frac{1}{t},\frac{1}{t},..., \frac{1}{t}]$. Next, we further analyze how different $Orn$ values affect the distance between RPSs.

\begin{figure}[htbp]
    \subfigure[\label{figure11a}The distance between $RPS_1$ and $RPS_2$ ($BPA_1$ and $BPA_2$)]{
       \centering
        \includegraphics[width=0.435\textwidth]{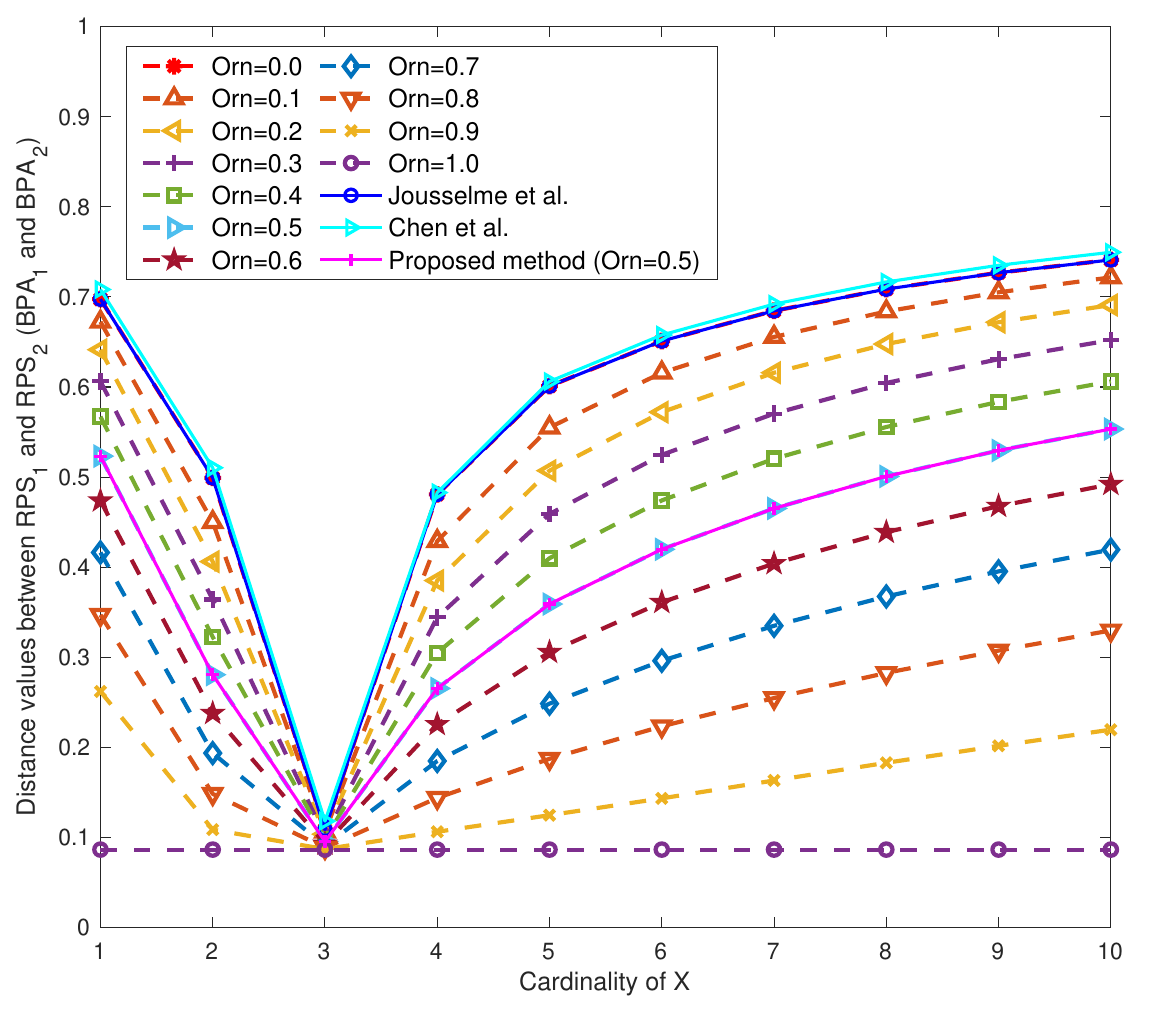}
    }
    \hspace{10mm}\subfigure[\label{figure11b}The distance between $RPS_1$ and $RPS_3$ ($BPA_1$ and $BPA_3$)]{
        \centering
        \includegraphics[width=0.45\textwidth]{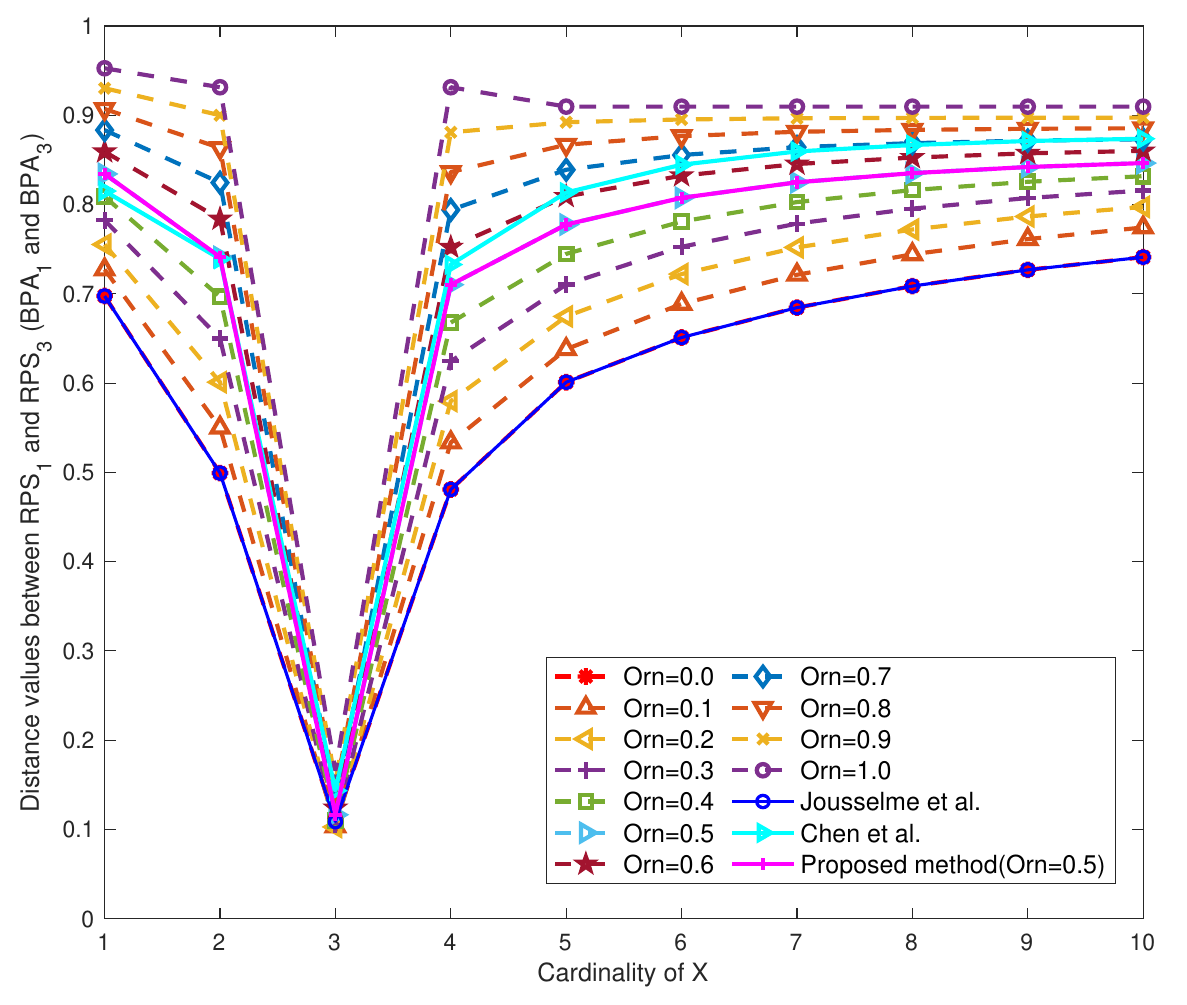}
    }\caption{The distance between RPSs (BPAs) when $Orn$ takes different values in Example 5.5}
    \label{Example4.5(Orn)}
\end{figure}

Figures \ref{figure11a} and \ref{figure11b} show the distances between the RPSs in Example 5.5 when parameter $Orn$ varies from 0 to 1 with a step of 0.1. The specific distance values are presented in Tables \ref{tab10} and \ref{tab11}. For better display and comparison,  the results obtained using Chen et al.'s method\cite{chen2023distance} and the Jousselme distance\cite{jousselme2001new} of the corresponding mass functions are still retained in Figure \ref{figure11a} and \ref{figure11b}.

\begin{table}[!htbp]\small
   	\begin{center}
         	\setlength{\tabcolsep}{0.5cm}
     		\caption{The distance between $RPS_1$ and $RPS_2$ when $Orn$ takes different values in Example 5.5}
    		\begin{tabular}{p{13em}p{2.5em}p{2.5em}p{2.5em}p{2.5em}p{2.5em}}
     			\toprule
                     &\multicolumn{5}{c}{$d_{RPS}(Perm_1, Perm_2)$}  \\
                     \cmidrule(lr){2-6}
     			 \ \ \ \ \ \ \ \ \ \ \ \ \ \ \ \ \ $X$ & $orn$=0.0 &$orn$=0.1 & $orn$=0.2 & $orn$=0.3 &$orn$=0.4\\
     			\midrule
    			\specialrule{0em}{1pt}{1pt}
     			$F_1^1: \tau_1$ & 0.6975   &0.6724& 0.6414& 0.6063&0.5671  \\
         		\specialrule{0em}{1pt}{1pt}
     			$F_3^1:\tau_1\tau_2$& 0.4992 &0.4501& 0.4062&0.3642&0.3226 \\
         		\specialrule{0em}{1pt}{1pt}
     			$F_7^1:\tau_1\tau_2\tau_3$&0.1088  &0.1066& 0.1037& 0.1009&0.0981   \\
         		\specialrule{0em}{1pt}{1pt}
     			$F_{15}^1:\tau_1\tau_2\tau_3\tau_4$&0.4808  &0.4290& 0.3852&0.3445 & 0.3049   \\
         		\specialrule{0em}{1pt}{1pt}
     			$F_{31}^1:\tau_1\tau_2\tau_3\tau_4\tau_5$&0.6010  & 0.5550& 0.5071& 0.4591&0.4100 \\
          		\specialrule{0em}{1pt}{1pt}
     			$F_{63}^1:\tau_1\tau_2\tau_3\tau_4\tau_5\tau_6$&0.6510  & 0.6159& 0.5721&0.5247& 0.4741  \\
          		\specialrule{0em}{1pt}{1pt}
     			$F_{127}^1:\tau_1\tau_2\tau_3\tau_4\tau_5\tau_6\tau_7$&0.6845  &  0.6556& 0.6159&0.5705&0.5204   \\
          		\specialrule{0em}{1pt}{1pt}
     			$F_{255}^1:\tau_1\tau_2\tau_3\tau_4\tau_5\tau_6\tau_7\tau_8$&0.7086  & 0.6839& 0.6477& 0.6045 &  0.5556 \\
          		\specialrule{0em}{1pt}{1pt}
     			$F_{511}^1:\tau_1\tau_2\tau_3\tau_4\tau_5\tau_6\tau_7\tau_8\tau_9$&0.7268  & 0.7051& 0.6719&0.6309& 0.5835   \\
          		\specialrule{0em}{1pt}{1pt}
     			$F_{1023}^1:\tau_1\tau_2\tau_3\tau_4\tau_5\tau_6\tau_7\tau_8\tau_9\tau_{10}$&0.7411  & 0.7217& 0.6910&0.6520&0.6062  \\
            		\toprule
                    &\multicolumn{5}{c}{$d_{RPS}(Perm_1, Perm_2)$}  \\
                     \cmidrule(lr){2-6}
     			 \ \ \ \ \ \ \ \ \ \ \ \ \ \ \ \ \ $X$ & $orn$=0.5 &$orn$=0.6 & $orn$=0.7 & $orn$=0.8 &$orn$=0.9\\
	             \midrule
     			\specialrule{0em}{1pt}{1pt}
     			$F_1^1: \tau_1$ & 0.5233& 0.4737& 0.4163& 0.3471& 0.2621    \\
         		\specialrule{0em}{1pt}{1pt}
     			$F_3^1:\tau_1\tau_2$ & 0.2807& 0.2379& 0.1937& 0.1485& 0.1088\\
         		\specialrule{0em}{1pt}{1pt}
     			$F_7^1:\tau_1\tau_2\tau_3$  &  0.0955& 0.0930& 0.0906& 0.0886& 0.0871    \\
                    \specialrule{0em}{1pt}{1pt}
       			$F_{15}^1:\tau_1\tau_2\tau_3\tau_4$  & 0.2655& 0.2256& 0.1849& 0.1440& 0.1064    \\
         		\specialrule{0em}{1pt}{1pt}
     			$F_{31}^1:\tau_1\tau_2\tau_3\tau_4\tau_5$  & 0.3591& 0.3056& 0.2486& 0.1874& 0.1248   \\
          		\specialrule{0em}{1pt}{1pt}
     			$F_{63}^1:\tau_1\tau_2\tau_3\tau_4\tau_5\tau_6$  &  0.4199& 0.3610& 0.2962& 0.2234& 0.1434    \\
          		\specialrule{0em}{1pt}{1pt}
     			$F_{127}^1:\tau_1\tau_2\tau_3\tau_4\tau_5\tau_6\tau_7$  & 0.4653& 0.4041& 0.3350& 0.2548& 0.1633   \\
          		\specialrule{0em}{1pt}{1pt}
     			$F_{255}^1:\tau_1\tau_2\tau_3\tau_4\tau_5\tau_6\tau_7\tau_8$  & 0.5008& 0.4389& 0.3676& 0.2827& 0.1829  \\
         		\specialrule{0em}{1pt}{1pt}
     			$F_{511}^1:\tau_1\tau_2\tau_3\tau_4\tau_5\tau_6\tau_7\tau_8\tau_9$  & 0.5296& 0.4677& 0.3954& 0.3075& 0.2018   \\
          		\specialrule{0em}{1pt}{1pt}
     			$F_{1023}^1:\tau_1\tau_2\tau_3\tau_4\tau_5\tau_6\tau_7\tau_8\tau_9\tau_{10}$  &0.5534& 0.4921& 0.4195& 0.3298& 0.2197 \\
    			\bottomrule
     			\label{tab10}
     		\end{tabular}
     	\end{center}
     \end{table}

\begin{table}[!htbp]\small
   	\begin{center}
         	\setlength{\tabcolsep}{0.5cm}
     		\caption{The distance between $RPS_1$ and $RPS_3$ when $Orn$ takes different values in Example 5.5}
    		\begin{tabular}{p{13.5em}p{2.5em}p{2.5em}p{2.5em}p{2.5em}p{2.5em}}
     			\toprule
                     &\multicolumn{5}{c}{$d_{RPS}(Perm_1, Perm_3)$}  \\
                     \cmidrule(lr){2-6}
     			 \ \ \ \ \ \ \ \ \ \ \ \ \ \ \ \ \ $X$ & $orn$=0.0 &$orn$=0.1 & $orn$=0.2 & $orn$=0.3 &$orn$=0.4\\
     			\midrule
    			\specialrule{0em}{1pt}{1pt}
    			$F_1^1: \tau_1$ & 0.6975 &0.7272  &0.7552 &0.7824  &0.8087  \\
        		\specialrule{0em}{1pt}{1pt}
    			$F_3^2:\tau_2\tau_1$ &0.4992  & 0.5501 &0.6010 &0.6500  & 0.6967\\
        		\specialrule{0em}{1pt}{1pt}
    			$F_7^6:\tau_3\tau_2\tau_1$  & 0.1088  & 0.1035 &0.1028  & 0.1053 &0.1101   \\
        		\specialrule{0em}{1pt}{1pt}
    			$F_{15}^{24}:\tau_4\tau_3\tau_2\tau_1$  & 0.4808 &0.5335  &0.5798& 0.6244 &  0.6677 \\
        		\specialrule{0em}{1pt}{1pt}
    			$F_{31}^{120}:\tau_5\tau_4\tau_3\tau_2\tau_1$  & 0.6010 & 0.6379 & 0.6747&  0.7104 & 0.7447 \\
         		\specialrule{0em}{1pt}{1pt}
    			$F_{63}^{720}:\tau_6\tau_5\tau_4\tau_3\tau_2\tau_1$  & 0.6510 &0.6887  &0.7221& 0.7526 &0.7810  \\
         		\specialrule{0em}{1pt}{1pt}
    			$F_{127}^{5040}:\tau_7\tau_6\tau_5\tau_4\tau_3\tau_2\tau_1$  & 0.6845  &0.7213 &0.7520 & 0.7786  &  0.8028 \\
         		\specialrule{0em}{1pt}{1pt}
    			$F_{255}^{40320}:\tau_8\tau_7\tau_6\tau_5\tau_4\tau_3\tau_2\tau_1$  & 0.7086  &0.7442 &0.7720 & 0.7954  & 0.8162 \\
         		\specialrule{0em}{1pt}{1pt}
    			$F_{511}^{362880}:\tau_9\tau_8\tau_7\tau_6\tau_5\tau_4\tau_3\tau_2\tau_1$  & 0.7268  & 0.7613 &0.7866  &0.8073 &0.8255 \\
         		\specialrule{0em}{1pt}{1pt}
    			$F_{1023}^{3628800}:\tau_{10}\tau_9\tau_8\tau_7\tau_6\tau_5\tau_4\tau_3\tau_2\tau_1$  &0.7411 &  0.7743  &0.7973 &0.8157  & 0.8318  \\
            		\toprule
                    &\multicolumn{5}{c}{$d_{RPS}(Perm_1, Perm_3)$}  \\
                     \cmidrule(lr){2-6}
     			 \ \ \ \ \ \ \ \ \ \ \ \ \ \ \ \ \ $X$ & $orn$=0.5 &$orn$=0.6 & $orn$=0.7 & $orn$=0.8 &$orn$=0.9\\
	             \midrule
    			\specialrule{0em}{1pt}{1pt}
    			$F_1^1: \tau_1$  &0.8343& 0.8592 & 0.8834 &0.9071 &0.9303  \\
        		\specialrule{0em}{1pt}{1pt}
    			$F_3^2:\tau_2\tau_1$  &0.7413&0.7838  &0.8245&0.8634&0.9000  \\
        		\specialrule{0em}{1pt}{1pt}
    			$F_7^6:\tau_3\tau_2\tau_1$  &0.1165& 0.1241 &0.1327&0.1425&0.1535  \\
        		\specialrule{0em}{1pt}{1pt}
    			$F_{15}^{24}:\tau_4\tau_3\tau_2\tau_1$  &0.7102& 0.7521 &0.7940&0.8364 & 0.8809  \\
        		\specialrule{0em}{1pt}{1pt}
    			$F_{31}^{120}:\tau_5\tau_4\tau_3\tau_2\tau_1$   &0.7776& 0.8090 &0.8389&0.8670&0.8922 \\
         		\specialrule{0em}{1pt}{1pt}
    			$F_{63}^{720}:\tau_6\tau_5\tau_4\tau_3\tau_2\tau_1$    &0.8075& 0.8323  &0.8555&0.8767&0.8953  \\
         		\specialrule{0em}{1pt}{1pt}
    			$F_{127}^{5040}:\tau_7\tau_6\tau_5\tau_4\tau_3\tau_2\tau_1$   &0.8249& 0.8454 & 0.8643&0.8815&0.8966 \\
         		\specialrule{0em}{1pt}{1pt}
    			$F_{255}^{40320}:\tau_8\tau_7\tau_6\tau_5\tau_4\tau_3\tau_2\tau_1$  &0.8351 & 0.8526 &0.8688&0.8836&0.8971  \\
         		\specialrule{0em}{1pt}{1pt}
    			$F_{511}^{362880}:\tau_9\tau_8\tau_7\tau_6\tau_5\tau_4\tau_3\tau_2\tau_1$  &0.8420 &0.8574  &0.8716&0.8849&0.8973  \\
         		\specialrule{0em}{1pt}{1pt}
    			$F_{1023}^{3628800}:\tau_{10}\tau_9\tau_8\tau_7\tau_6\tau_5\tau_4\tau_3\tau_2\tau_1$   &0.8464 & 0.8601 &0.8731 &0.8854  & 0.8973 \\
    			\bottomrule
     			\label{tab11}
     		\end{tabular}
     	\end{center}
     \end{table}

When the $Orn=0$, the weight vector is given by $\mathbf{w}^{[0]}=[0, 0, ..., 1]$, with the entire weight concentrated on the deepest layer, corresponding to the permutation itself. In this case, the cumulative Jaccard index degenerates into the classical Jaccard index, and the proposed RPS distance measure method degenerates into the Jousselme distance between the corresponding BPAs. Consequently, we can observe from Figure \ref{figure11a} and \ref{figure11b} that the results between RPSs when $Orn=0$ coincide with those of the Jousselme distance computed between the corresponding BPAs.

When $Orn\in [0,0.5)$, the weight vector $\mathbf{w}^{[Orn]}$ forms a non-decreasing sequence, with weights gradually increasing from top to bottom. Moreover, as $Orn$ decreases, the bottom weights increase, progressively weakening the top-weightiness property. Since $X\in\mathcal{OF}_1$ affects the distance between $RPS_1$ and $RPS_2$ by continuously adding low-ranking elements, an increase in the weights assigned to the bottom position results in a larger distance between $RPS_1$ and $RPS_2$.  That is, the smaller the value of $Orn$, the larger the distance between $RPS_1$ and $RPS_2$, as shown in Figure \ref{figure11a} and Table \ref{tab10}. When $Orn\in (0.5,1]$, $\mathbf{w}^{[Orn]}$ is a non-increasing sequence, the weights gradually decrease from top to bottom, while the top weights increase as $Orn$ increases, further enhancing the top-weightiness property. The consistency in qualitative propensity between the top-ranked elements in $X\in\mathcal{OF}_1$ and $F_{7}^{1}\in\mathcal{OF}_2$ consequently leads to a decrease in the distance between $RPS_1$ and $RPS_2$ as $Orn$ increases. Until $Orn$ takes the value of 1, the weight vector is $\mathbf{w}^{[1]}=[1, 0, ..., 0]$, which means that only the top one focus element in the ordered focal set is considered. At this point, the behavior of $X$ continuously adding lower-ranked elements no longer affects the distance between $RPS_1$ and $RPS_2$, and the distance reaches its minimum value of 0.0866.

However, for the distance between $RPS_1$ and $RPS_3$ in Example 5.5, an opposite trend is observed, as shown in Figure \ref{figure11b} and Table \ref{tab11}: the distance between $RPS_1$ and $RPS_3$ increases as the value of $Orn$ increases. Differently, when calculating the distance between $RPS_1$ and $RPS_3$, $X$ changes from $\tau_1$ to $\tau_{10}\tau_9...\tau_1$ in reverse order. At this time, changes in the cardinality of $X$ increase the inconsistency among the top-ranked elements in the ordered focal set. Therefore, the increase in top weights leads to a larger distance between $RPS_1$ and $RPS_3$, resulting in a trend opposite to that observed between $RPS_1$ and $RPS_2$.

Compared with the Chen et al.' method\cite{chen2023distance}, the proposed RPS distance measure provides decision-makers with more flexible choices. Note that the parameter $Orn$ does not alter the inherent top-weightiness property of the proposed method but merely offers the option to weaken or reinforce this characteristic. Unless otherwise specified, $Orn$ defaults to 0.5, reflecting a neutral stance on the part of the decision-maker. If the decision-maker is more concerned with the top-ranked focal elements, the top-weightiness property can be enhanced by setting a larger value of $Orn$; conversely, if the decision-maker pays more attention to the lower-ranked focal elements, a smaller $Orn$ can be used to weaken this property. When the decision-maker does not require the top-weightiness property, $Orn$ can be set to 0, in which case the proposed method degenerates into the Jousselme distance\cite{jousselme2001new}. This indicates that the order information in the ordered focal set is no longer considered, and all focal elements are treated equally.

\section{conclusion}\label{conclusion}
This paper discusses the distance between two information distributions represented by layer-2 belief structure. First, we examine the differing behaviors and performances of the distance measure from the perspectives of RFS and TBM. Under the TBM framework, the RPS' layer-2 belief structure interprets the order introduced in the focal set as qualitative propensity information that indicates the decision-maker belief transfer propensity, characterized by higher-ranked elements being considered more important. Specifically targeting this characteristic, we introduce the new concept of cumulative Jaccard index to measure the similarity between two permutations. Then, we use it to construct the weighting matrix in the $L_2$ distance and further propose an RPS distance measure method based on the cumulative Jaccard index matrix. Additionally, we analyze the properties of the proposed method from both the metric and structural aspects, including an examination of the positive definiteness of the cumulative Jaccard index matrix and the corresponding correction scheme. Furthermore, we compare the proposed method with the existing one\cite{chen2023distance} through some numerical examples. The results of comparative experiments reveal the following advantages of the proposed method. (1) The proposed method has a natural top-weightiness property. Simply put, inconsistency between higher-ranked elements will lead to larger distances between RPSs, which is consistent with the RPS' layer-2 belief structure interpretation view. A parameter $Orn$ is provided to allow the decision-maker to enhance or weaken this property based on individual preferences or specific application scenarios. It should be pointed out that when $Orn$ is set to 0, the proposed method degenerates into the Jousselme distance\cite{jousselme2001new} in DST. (2) The proposed method has the ability to measure the distance between RPSs at any depth. The setting of parameter $t$ allows decision-makers to focus only on what they want to focus on and ignore redundancy. (3) The proposed method overcomes the limitations of the existing distance measure proposed by Chen et al.\cite{chen2023distance}, which fails to adhere to any interpretation of information distribution and does not satisfy axiomatic properties.
\section*{CRediT authorship contribution statement}
\textbf{Ruolan Cheng}: Conceptualization, Methodology, Formal analysis, Validation, Investigation, Writing - original draft, Writing - review \& editing. \textbf{Yong Deng}: Resources, Supervision, Funding acquisition, Writing - review \& editing. \textbf{Serafín Moral}: Formal analysis, Validation, Writing - review \& editing. \textbf{José Ramón Trillo}: Conceptualization, Methodology, Formal analysis, Validation, Resources, Funding acquisition, Writing - review \& editing.

\section*{Declaration of Competing Interest}
The authors declare that they have no known competing financial interests or personal relationships that could have appeared to influence the work reported in this paper.

\section*{Acknowledgment}
This work is supported by the National Natural Science Foundation of China (Grant No. 62373078), China Scholarship Council (202306070060), and the grant PID2022-139297OB-I00 funded by MCIN / AEI/ 10.13039/501100011033 and by "ERDF A way of making Europe". Moreover, it is part of the project C-ING-165-UGR23, co-funded by the Regional Ministry of University, Research and Innovation and by the European Union under the Andalusia ERDF Program 2021-2027.

\bibliography{References}
\bibliographystyle{elsarticle-num}

\end{document}